\documentclass{article}
\PassOptionsToPackage{numbers, compress}{natbib}

\usepackage[preprint]{style}

\usepackage[utf8]{inputenc} 
\usepackage[T1]{fontenc}    
\usepackage{hyperref}       
\usepackage{url}            
\usepackage{booktabs}       
\usepackage{amsfonts}       
\usepackage{nicefrac}       
\usepackage{microtype}      
\usepackage{xcolor}         
\usepackage{colortbl}
\usepackage{graphicx}
\usepackage{amsmath}
\usepackage{amsmath}
\usepackage{amssymb}
\usepackage{algorithm}
\usepackage{algorithmicx}
\usepackage{algpseudocode}

\usepackage{wrapfig}
\usepackage{tikz-cd}
\usepackage{graphicx}
\usepackage{amsmath}
\usepackage{amssymb}
\usepackage{amsthm}
\usepackage{enumitem}
\usepackage{algorithm}
\usepackage{algorithmicx}
\usepackage{algpseudocode}
\usepackage{wrapfig}

\usepackage{booktabs}
\usepackage{tabularx}
\usepackage{multirow}
\usepackage{siunitx}
\usepackage{algorithm}
\usepackage{algpseudocode}

\title{DFVEdit: Conditional Delta Flow Vector \\for Zero-shot Video Editing}

\author{
  Lingling Cai$^{1}$ \quad Kang Zhao$^{2}$\quad Hangjie Yuan${^{3,1}}$ \quad Xiang Wang$^{2,4}$\quad\\
  \textbf{Yingya Zhang$^{2}$} \quad \textbf{Kejie Huang$^{1}$}\\
  \\
  $^1$Zhejiang University \quad  $^2$Tongyi Lab\quad \\$^3$DAMO Academy\quad $^4$Huazhong University of Science and Technology\\
  \texttt{cailingling22@zju.edu.cn, zhaokang.zk@alibaba-inc.com, hj.yuan@zju.edu.cn,}\\
  \texttt{xiaolao.wx@alibaba-inc.com, yingya.zyy@alibaba-inc.com, huangkejie@zju.edu.cn}\\
  \\
\url{https://dfvedit.github.io}
}

\begin{document}
\maketitle
\begin{abstract}
The advent of Video Diffusion Transformers (Video DiTs) marks a milestone in video generation. However, directly applying existing video editing methods to Video DiTs often incurs substantial computational overhead, due to resource-intensive attention modification or finetuning. 
To alleviate this problem, we present DFVEdit, an efficient zero-shot video editing method tailored for Video DiTs. DFVEdit eliminates the need for both attention modification  and fine-tuning by directly operating on clean latents via flow transformation. To be more specific, we observe that editing and sampling can be unified under the continuous flow perspective. Building upon this foundation, we propose the Conditional Delta Flow Vector (CDFV) -- a theoretically unbiased estimation of DFV -- and integrate Implicit Cross Attention (ICA) guidance as well as Embedding Reinforcement (ER) to further enhance editing quality. DFVEdit excels in practical efficiency, offering at least 20x inference speed-up and 85\% memory reduction on Video DiTs compared to attention-engineering-based editing methods. Extensive quantitative and qualitative experiments demonstrate that DFVEdit can be seamlessly applied to popular Video DiTs (\emph{e.g.}, CogVideoX and Wan2.1), attaining state-of-the-art performance on structural fidelity, spatial-temporal consistency, and editing quality.
\end{abstract}

\section{Introduction}
\label{sec:intro}
\begin{wrapfigure}{r}{0.45\textwidth} 
  \label{fig:insight1}
  \centering
    \vspace{-4em}
  \includegraphics[width=0.45\textwidth]{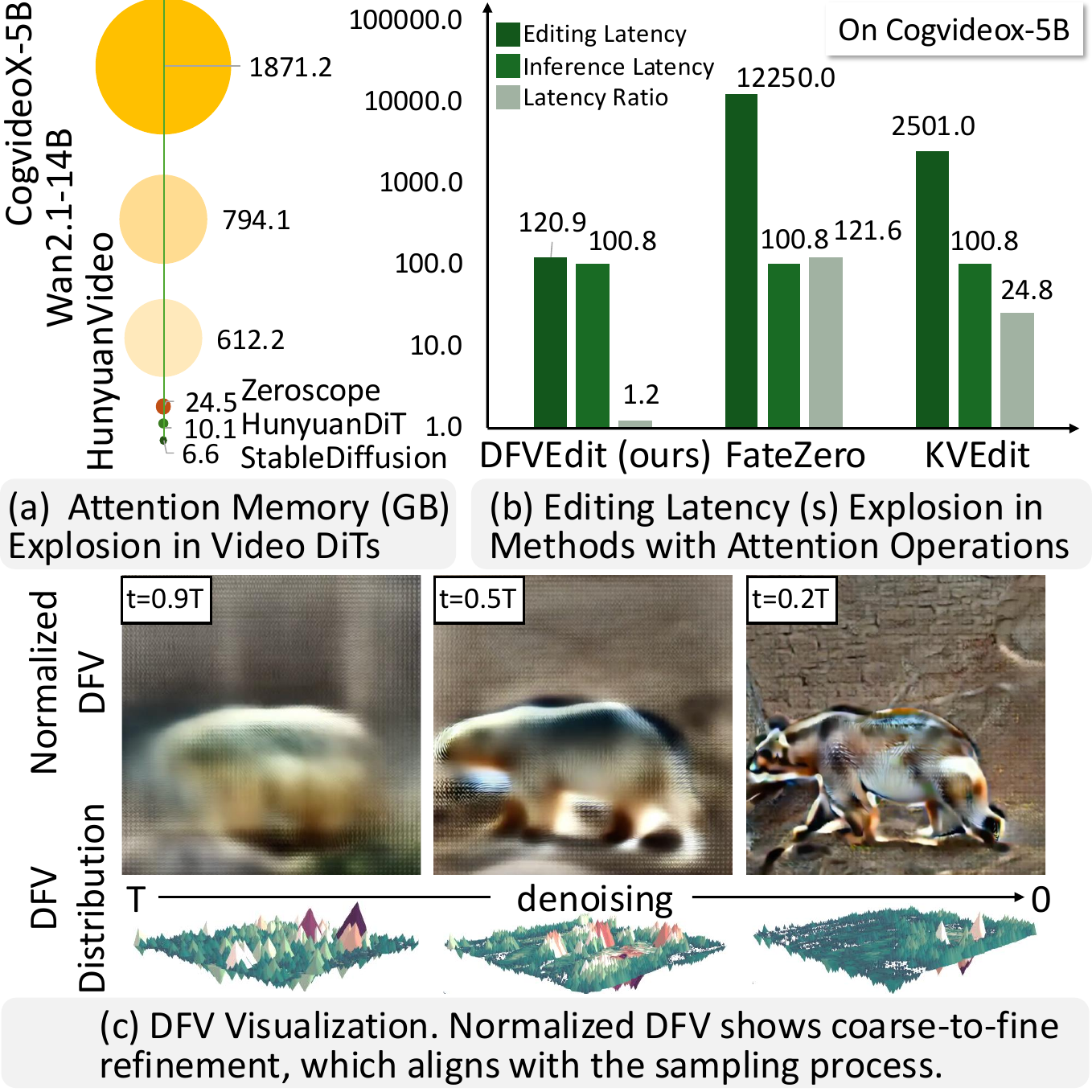} 
      \vspace{-1em}
  \caption{Key insight and motivation.}
    \vspace{-4em}

\end{wrapfigure}

In the wave of digitization, video creation has become a dominant form of entertainment. In response, research on controllable video generation holds considerable practical importance. While Video Diffusion Transformer (DiT) models~\cite{yang2024cogvideox,kong2024hunyuanvideo,peebles2023scalable,wang2025wan} have revolutionized video synthesis quality, and DiT-based image editing methods~\cite{feng2024dit4edit,kulikov2024flowedit,zhu2025kv,dalva2024fluxspace,rout2024semantic,jiao2025uniedit} have achieved remarkable success, video editing remains challenging in preserving spatiotemporal fidelity. Critically, existing video editing methods do not fully exploit the capabilities of Video DiTs, limiting the potential for high-quality controllable video generation.

Existing video editing techniques mainly follow two paradigms: training-based methods~\cite{singer2022make, wu2023tune, shin2024edit, liu2024videop2p} and zero-shot methods~\cite{qi2023fatezero, cai2024freemask, geyer2023tokenflow, zhang2023controlvideo, yangvideograin, wang2024videodirector}. Given that the former requires resource-intensive finetuning, our work focuses on training-free video editing. For training-free video editing, a high-quality pre-trained base model is crucial. Early video editing methods primarily utilized image diffusion models~\cite{rombach2022high, song2020denoising}, which suffered from temporal inconsistencies due to the lack of capable video diffusion models. These early methods~\cite{khachatryan2023text2video, qi2023fatezero, geyer2023tokenflow, yatim2024space} not only had to ensure structural integrity and editing accuracy but also required significant effort to enhance temporal coherence. In contrast, methods~\cite{cai2024freemask, ku2024anyv2v} based on video diffusion models naturally excel in temporal consistency, leading us to leverage the latest Video DiTs~\cite{wang2025wan, yang2024cogvideox, kong2024hunyuanvideo} for video editing. Regardless of the type of base models, achieving high fidelity and temporal consistency hinges on attention engineering in most existing methods, including various attention caching and modification techniques. The key to effective attention engineering is that attentions (including keys, queries, and values) contain the spatial-temporal information of the source video, allowing for smooth editing of target regions while preserving the original content's integrity. However, attention mechanisms now consume hundreds of gigabytes of memory (Fig.~\ref{fig:insight1}(a)) in Video DiTs~\cite{yang2024cogvideox, wang2025wan, kong2024hunyuanvideo}, a significant increase from previous usage in Unet-based diffusion models~\cite{wang2023modelscope, rombach2022high, song2020denoising} and image DiT models~\cite{li2024hunyuan, yang20241} at the gigabyte scale. This suggests that traditional attention engineering techniques are incompatible with Video DiTs, creating an urgent need for methods that preserve editing quality while improving computational efficiency.

Motivated by this inefficiency, we shift the focus from attention to input latents and introduce a continuous flow transformation framework for direct video latent refinement. We observe that the standard sampling process in video diffusion models—whether based on Score Matching~\cite{song2020score} or Flow Matching~\cite{lipman2022flow}—can be unified under a continuous flow perspective. Based on this insight, we demonstrate that editing from the source to the target video naturally forms a time-dependent flow vector field (Fig.~\ref{fig:insight1}(c)), which we term the Delta Flow Vector (DFV).

Building upon this foundation, we introduce the Conditional Delta Flow Vector (CDFV) to estimate the flow from source to target latent, incorporating Implicit Cross Attention Guidance (ICA) and Embedding Reinforcement (ER) to further improve editing accuracy. The CDFV in Video DiTs inherently enforces spatial-temporal dependencies while its divergence directly determines update weights. This physically grounded formulation provides two fundamental advantages over approximation-based latent-refinement approaches like DDS~\cite{hertz2023delta} and SDS~\cite{pooledreamfusion}: (1) \textit{theoretical unification} by modeling both sampling and editing from the continuous flow perspective and 
(2) \textit{computational efficiency} through divergence-determined and hyperparameter-free weights that eliminate heuristic scheduling and overcome low convergence issues inherent to shallow approximations. 
Moreover, for the seamless application to video editing, we enhanced spatiotemporal coherence by intrinsically avoiding randomness bias while incorporating ICA guidance and ER mechanisms (Fig.~\ref{fig: ablation1}). 
Experiments show DFVEdit achieves at least 20× speed-up and 85$\%$ memory reduction over attention-engineering-based methods on Video DiTs (\emph{e.g.}, CogVideoX, Wan2.1), while maintaining SOTA performance in fidelity, temporal consistency, and editing quality. Consequently, our approach offers an efficient and versatile solution for zero-shot video editing on Video DiTs.

\section{Related work}
\label{related: video_editing}
\textbf{Video Diffusion Transformer.}
Video Diffusion Transformers have evolved from early 3D-UNet-based designs~\cite{zhang2023i2vgen,wang2023modelscope,blattmann2023stable,chen2024videocrafter2} to modern 3D-Transformer-based designs~\cite{peebles2023scalable}. Advanced models such as Open-Sora~\cite{zheng2024open,lin2024open}, CogVideoX~\cite{yang2024cogvideox}, HunyuanVideo~\cite{kong2024hunyuanvideo} and Wan~\cite{wang2025wan} have all or part of the following key innovations: replacement of 3D-UNets with scalable 3D-Transformer blocks; integration of cross-attention and self-attention into a unified 3D-full-attention~\cite{yang2024cogvideox,kong2024hunyuanvideo}; and adoption of 3D-VAE~\cite{yang2024cogvideox} for spatiotemporal latent compression. Some Video DiTs~\cite{li2024hunyuan,wang2025wan} are combined with Flow Matching~\cite{lipman2022flow} while others~\cite{yang2024cogvideox} adopt SDE~\cite{song2020score} samplers like DPM-solver~\cite{lu2022dpm}.

\textbf{Image editing on Diffusion Transformer.}
With the rise of Diffusion Transformer~\cite{peebles2023scalable}, DiT-based image editing methods~\cite{yang20241,li2024hunyuan} have emerged.
However, directly applying image editing methods to videos often fails to address temporal consistency and motion fidelity. Additionally, adapting them to Video DiTs introduces extra challenges.
Firstly, generalization limitations occur when applying methods~\cite{dalva2024fluxspace,kulikov2024flowedit,rout2024semantic,jiao2025uniedit,garibi2024renoise,deutch2024turboedit} that rely on rectified flow~\cite{esser2024scaling} or distilled few-step models~\cite{sauer2024adversarial} to Video DiTs that are not combined with rectified flow or distillation techniques.  Secondly, efficiency limitations are present for image editing methods~\cite{nguyen2024swiftedit} that require finetuning.  Furthermore, even generalized and efficient methods like DiT4Edit~\cite{feng2024dit4edit} and KVEdit~\cite{zhu2025kv}, which use attention or key-value caching and modification, still face prohibitive computational costs due to the more massive attention overhead in Video DiTs compared to image DiTs.
\begin{figure*}[htbp] 
  \centering 
  \includegraphics[width=1\textwidth]{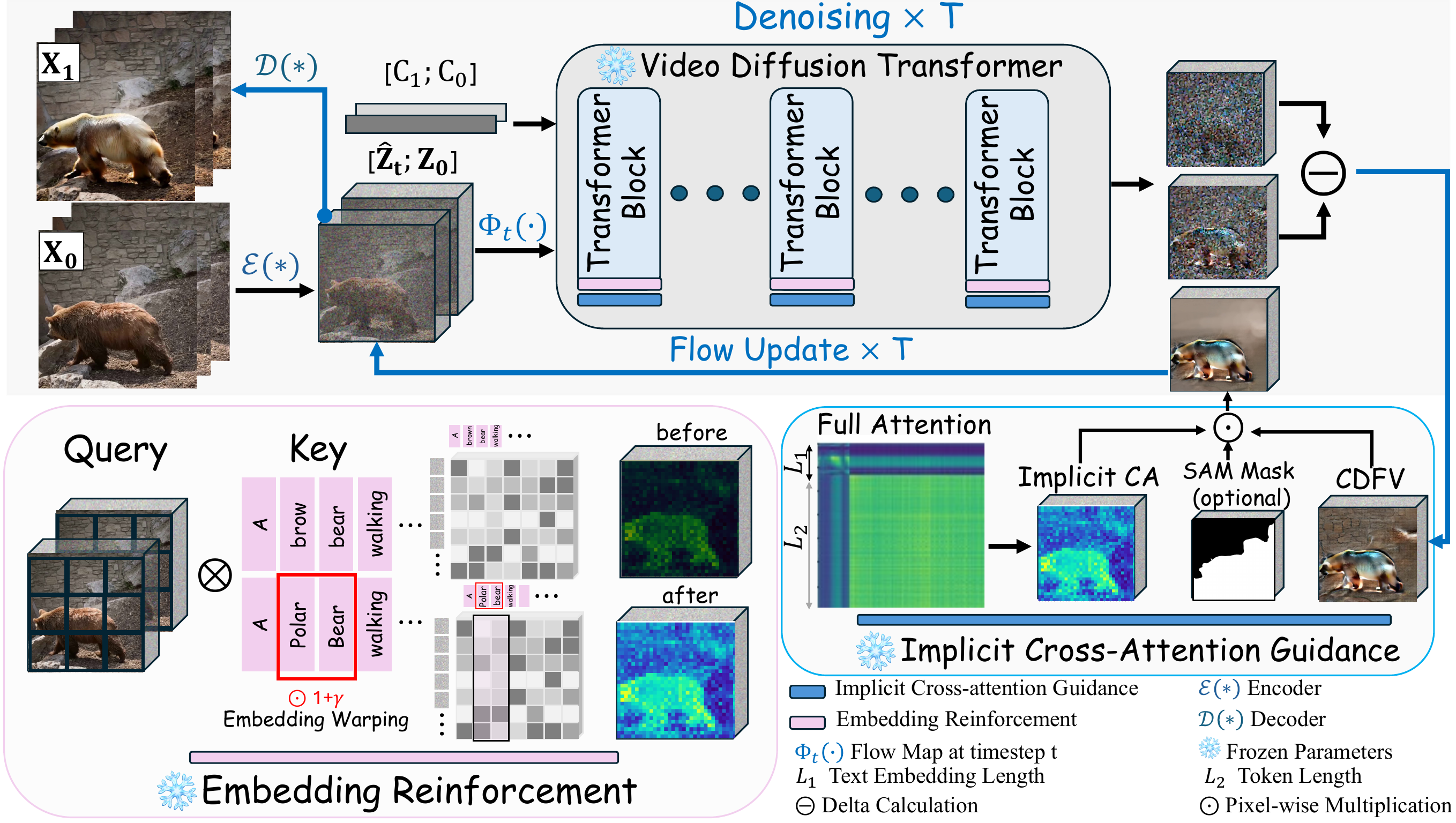} 

\caption{\textbf{DFVEdit overview.} Follow these steps for DFVEdit:
(1) Encode $\mathbf{X}_0$ into the latent space $\mathbf{Z}_0$, and initialize the target latent variable as $\hat{\mathbf{Z}}_T = \mathbf{Z}_0$. 
(2) Transform $[\hat{\mathbf{Z}}_T; \mathbf{Z}_0]$ via the flow map $\Phi_T(\cdot)$. 
(3) Feed the result with prompt embeddings $[C_1, C_0]$ into the Video Diffusion Transformer, compute the delta difference to obtain the CDFV at timestep $T$, then refine it using ER and ICA. 
(4) Update $\hat{\mathbf{Z}}_T \rightarrow \hat{\mathbf{Z}}_{T-1}$ using the enhanced CDFV, and iterate (1)-(4) until reaching $\hat{\mathbf{Z}}_0$. 
(5) Decode $\hat{\mathbf{Z}}_0$ to generate the target video $\mathbf{X}_1$.}
  \label{fig: framework} 
  \vspace{-1em}
\end{figure*}

\textbf{Video editing.} 
Video editing via diffusion models is dominated by two paradigms: training-based and training-free methods. Training-based approaches \cite{jiang2025vace, peruzzo2024vase, esser2023structure, gu2024videoswap, zi2025cococo, wang2024videocomposer, wu2023tune, liu2024videop2p} enhance pre-trained image diffusion models \cite{rombach2022high} with spatiotemporal modules, optimizing for complex edits but at high computational costs, limiting real-time applications. Conversely, training-free methods emphasize computational efficiency and real-time capability.
Training-free video editing commonly involves two stages: latent space initialization and editing condition injection. Latent space initialization typically follows three paradigms: (1) forward diffusion with some steps for preserving low-frequency features~\cite{meng2021sdedit,yang2023rerender}, (2) DDIM~\cite{song2020denoising} inversion for enabling deterministic reconstruction~\cite{qi2023fatezero,geyer2023tokenflow}, or (3) direct source latent usage~\cite{hertz2023delta,pooledreamfusion}. 
For editing condition injection, most existing zero-shot methods heavily rely on \textit{attention engineering} to maintain spatial-temporal fidelity. For instance, FateZero~\cite{qi2023fatezero} enhances temporal consistency by caching attention maps from DDIM~\cite{song2020denoising} inversion and integrating them into the denoising process; TokenFlow~\cite{geyer2023tokenflow} improves spatiotemporal coherence by leveraging cached attention outputs from DDIM inversion for inter-frame correspondences and incorporating extended attention blocks during denoising; VideoDirector~\cite{wang2024videodirector} achieves fine-grained editing via SAM~\cite{kirillov2023segment} masks by fusing self-attention with reconstruction attention and mask guidance; and VideoGrain~\cite{yangvideograin} realizes complex semantic structure modifications through SAM masks while operating on complex attention map modifications.
These attention-engineered methods face scalability challenges in Transformer blocks~\cite{vaswani2017attention}, particularly for Video DiTs~\cite{kong2024hunyuanvideo,wang2025wan} where attention memory demands grow dramatically (Fig.~\ref{fig:insight1}). 
Moreover, approaches~\cite{yoon2024frag,liu2024stablev2v,kuanyv2v,yatim2024space}  free of attention engineering suffer from structural degradation: FRAG~\cite{yoon2024frag} mitigates blurring and flickering through frequency processing but compromises fidelity due to basic DDIM inversion~\cite{songdenoising} for source content retention; DMT~\cite{yatim2024space} employs SSM~\cite{yatim2024space}  loss for motion transfer yet underperforms in detail preservation; and first-frame propagation methods (\emph{e.g.}, StableV2V~\cite{liu2024stablev2v}, AnyV2V~\cite{kuanyv2v}) introduce accumulating artifacts without full-frame coordination. In conclusion, designing efficient and high-quality editing methods tailored for Video DiTs remains a critical challenge.


\vspace{-1em}
\section{Method}
\vspace{-0.5em}
Fig.~\ref{fig: framework} provides an overview of DFVEdit. Given a source video $\mathbf{X_0} \in \mathbb{R}^{F \times 3 \times H \times W}$ comprising $F$ RGB frames at resolution $H \times W$, together with source and target text prompts $(P_0, P_1)$, our method supports both global stylization and local modifications (shape and attribute editing). The edited video $\mathbf{X_1}$ preserves spatiotemporal integrity in unedited regions while ensuring motion fidelity and precise alignment with $P_1$. Our approach leverages two key insights: manipulating latent space is more computationally efficient than manipulating attention (Fig.~\ref{fig:insight1}), and editing can be modeled as the continuous flow transformation between the source and target videos (Sec~\ref {subsec: rethinking}). We introduce the Conditional Delta Flow Vector (CDFV) (Sec~\ref {subsec: CDFV}) for this transformation. To enhance video editing performance, we utilize Implicit Cross-Attention Guidance and Embedding Enforcement (Sec~\ref{subsec: improvement}) to improve spatiotemporal fidelity.

\subsection{Unified continuous flow perspective on sampling and editing}
\label{subsec: rethinking}

Diffusion models include inverse and forward processes. The inverse process is typically parameterized as a Markov chain with learned Gaussian transitions, mapping noisy inputs to clean outputs. Conversely, the forward process gradually adds Gaussian noise to the clean input according to a variance schedule. As mentioned in~\cite{song2020improved, song2019generative,song2020score}, given a data input $x$, both inverse and forward processes can be regarded as overdamped Langevin Dynamics~\cite{uhlenbeck1930theory} (named Stochastic Differential Equation (SDE) in Score Matching~\cite{song2020score}):
\begin{align}
dx_t=f(x_t,t)dt+g(x_t,t)dW
\label{eq: sde}
\end{align}
where $f(x_t,t)$ is the drift coefficient corresponds to deterministic direction and $g(x_t,t)$ is the diffusion coefficient corresponds to disturbing intensity and $dW$ is a Wiener process and the probability density $P(x_t,t)$ can be described by introducing the Fokker-Planck equation~\cite{jordan1998variational} combined with the Ito's lemma~\cite{kloeden1992stochastic} and the concept of probability flow:
\begin{align}
\label{eq: sde2}
\frac{\partial P(x_t,t)}{\partial t} = -\nabla\left[\left(f(x_t,t)-\frac{g^2(x_t,t)}{2}\nabla logP(x_t,t)\right)P(x_t,t)\right]
\end{align}
Eq.~\ref{eq: sde2} generalizes traditional sampling methods like DDPM~\cite{ho2020denoising} and DDIM~\cite{song2020denoising}. \textbf{\textit{This formulation reveals that methods based on SDE~\cite{song2020score} obey the continuity equation principle of Flow Matching~\cite{lipman2022flow} and can be unified under a continuous flow perspective.}} The continuous flow is characterized by a vector field  $v_t(x_t)=f(x_t,t)-\frac{g^2(x_t,t)}{2}\nabla logP(x_t,t)$, enabling state transitions from $x_t$ to $x_{t+\Delta t}$ either through flow map $\Phi_t$ in Eq.~\ref{eq:7} or through its Euler discretized approximation in Eq.~\ref{eq:8}:
\begin{equation}
\left\{
\begin{aligned}
    & \frac{d}{dt}\Phi_t(x)=v_t(\Phi_t(x))\\
    & \Phi_0(x)=x    
\end{aligned}
\right.
\label{eq:7}
\end{equation}
\begin{align}
x_{t+\Delta t}=x_t+\Delta t*v_t(\Phi_t(x))
\label{eq:8}
\end{align}
As discussed in Section~\ref{related: video_editing}, zero-shot video editing includes two stages: latent space initialization and editing condition injection. The first stage involves a standard sampling process. In the second stage, we derive an isomorphism with sampling process by formulating video editing as: 
\begin{equation}
    X_{t-1}^{\text{edit}} = g_{\theta_{2,t}}\Big( X_t^{\text{edit}},\ \underbrace{\epsilon_{\theta_1}(X_t^{\text{edit}}, t)}_{\text{Canonical Denoiser}} + \lambda \underbrace{C(X_t^{\text{edit}}, t,*)}_{\text{Control Term}} \Big)
\label{eq: edit}
\end{equation}
where $\{X_t^{\text{edit}}\}_{t=0}^T$ defines the state trajectory of the edited video in the sampling process; $g_{\theta_{2,t}}$ is differentiable transition function parameterized by learnable $\theta_2$; $\epsilon_{\theta_1}$ is pretrained diffusion model with frozen $\theta_1$;
$C(x, t,*)$ is the control term with intensity $\lambda \geq 0$ and optional extra input $*$ . 
Under the Euler discretization scheme with step size $\Delta t \to 0$ and $\theta_2=\mathcal{I}$, the discrete process in Eq.~\ref{eq: edit} converges to the controlled SDE:
\begin{align}
    dX^{\text{edit}}_t = \underbrace{
  \left[ 
    -\frac{\beta(t)}{2} X^{\text{edit}}_t  + \frac{\beta(t)}{2} \nabla\log p_t(X^{\text{edit}}_t ) + \lambda \frac{\beta(t)}{2} \sigma(t) C(X^{\text{edit}}_t ,t,*)
  \right]
}_{f_{\theta_1}(X^{\text{edit}}_t ,t)} dt + \underbrace{\sqrt{\beta(t)}}_{g(t)} dW
\label{eq: edit2}
\end{align}
where $\nabla\log p_t(X^{\text{edit}}_t )$ is the score function, and $\sigma(t) = \sqrt{(1-\alpha(t))/\alpha(t)}$ is the signal-to-noise ratio coefficient with $\alpha(t) = e^{-\int_0^t \beta(s)ds}$. \textbf{\textit{The structural isomorphism between Eq.~\ref{eq: edit2} and the stochastic differential equation in Eq.~\ref{eq: sde} indicates that video editing processes can be represented within a continuous flow sampling framework}}, as shown in Eq.~\ref{eq:7} (see Appendix for more details).

\subsection{Conditional Delta Flow Vector}
\label{subsec: CDFV}
Building upon the isomorphic correspondence between editing and sampling, we introduce the Conditional Delta Flow Vector (CDFV) to establish a direct continuous flow bridge from the source video to the target video.

\noindent\textbf{Delta Flow Vector.}  
Given the initial distribution $p(Z_T) = \mathcal{N}(Z_T; 0, I)$ for the reverse process and a clean video latent $Z$, Eq.~\ref{eq:7} implies the existence of a time-dependent flow map $\Phi$ that:
\begin{equation}
\label{eq: 8}
Z=Z_T-\sum_{t=0}^{T}\Delta tv_t(\Phi_t(Z))
\end{equation} 
Assuming the source and target latents $(\hat{Z}_0,Z_0)$ and their corresponding prompts $(P_1, P_0)$ are given, we replace $Z$ in Eq.~\ref{eq: 8} with $Z_0$ and $\hat{Z}_0$ respectively and define the Delta Flow Vector (DFV) as $\Delta v_t(\hat{Z}_0, Z_0) = v_t(\Phi_t(\hat{Z}_0)) - v_t(\Phi_t(Z_0))$, and the target latent $\hat{Z}_0$ can be expressed in terms of the source latent $Z_0$ as:
\begin{equation}
\label{eq:9}
\hat{Z}_0 = Z_0 - \sum_{t=0}^{T} \Delta t \, \Delta v_t(\hat{Z}_0, Z_0).
\end{equation}
\textbf{\textit{Eq.~\ref{eq:9} establishes a continuous flow directly from the source latent $Z_0$ to the target latent $\hat{Z}_0$}}, with the vector field defined as $v_t=\Delta v_t(\hat{Z}_0, Z_0)$. While prior works~\cite{han2024proxedit,couairon2022diffedit,hertz2023delta} heuristically observed that latent differences indicate editing regions, we rigorously prove this as a special case of DFV when the transformation state and vector field satisfy the continuity equation (Eq.~\ref{eq:7}).

\noindent\textbf{Conditional Delta Flow Vector.}
The direct computation of $\Delta v_t(\hat{Z}_0, Z_0)$ is intractable since $\hat{Z}_0$ is the editing target. To resolve this problem, \textbf{\textit{we leverage the terminal condition of diffusion processes to derive an unbiased estimation of DFV.}} From Eq.~\ref{eq: sde2} we obtain $v_t(x_t)=f(x_t,t)-\frac{g^2(x_t,t)}{2}\nabla logP(x_t,t)$. As $t$ approaches $T$, and given that $P(x_t,t)$ is the probability density of $x_t$, if we set winner process of $Z_0$ and $\hat Z_0$ is equal, then $g(Z_0,t)=g(\hat Z_0,t)$. Consequently, as $t \to T$, both $P(Z_0,t)$ and $P(\hat{Z}_0,t)$ follow a normal distribution $\mathcal{N}(Z_T; 0, I)$ with zero mean and unit variance. Moreover, $\hat{Z}_t$ is equivalent to $Z_t$ as $t \to T$, and we have:
\begin{equation}
\label{eq: dfv}
\Delta v_t(\hat Z_0, Z_0)\underset{t \to T}=f_{\theta_1,c_1}(Z_t,t)-f_{\theta_1,c_0}(Z_t,t)
\end{equation} 
The latent $\hat{Z}_{T-\Delta t}$ can be updated using Eq.~\ref{eq:z_t}, which corresponds to applying the continuous flow map from $\hat{Z}_0$ as defined in Eq.~\ref{eq: dfv2}:
\begin{equation}
\label{eq:z_t}
\hat{Z}_{T-\Delta t} = Z_{T-\Delta t} - \Delta t \left[ f_{\theta,c_1}(Z_T, t) - f_{\theta,c_0}(Z_T, t) \right],
\end{equation}
\begin{equation}
\label{eq: dfv2}
\hat{Z}_{T-\Delta t} = \Phi_{T-\Delta t}(\hat{Z}_0).
\end{equation}
We sequentially obtain all $v_t(\Phi(\hat{Z}_0))$ and define the Conditional Delta Flow Vector (CDFV) in Eq.~\ref{eq:cdfv}. 
\begin{equation}
\label{eq:cdfv}
\left\{
\begin{aligned}
& \Delta v_t(Z_0,c_0,c_1)=v_{t,c_1}(\hat Z_t)-v_{t,c_0}(\Phi_t(Z_0)) \\
& \hat Z_{T}=\Phi_T(Z_0)
\end{aligned}
\right.
\end{equation}
Theoretically, the CDFV provides an unbiased estimate of DFV. By using the CDFV as a control term, defined in Eq.~\ref{eq:control}, and integrating it into Eq.~\ref{eq: edit2}, we maintain a computational complexity similar to that of the basic sampling process. See the Appendix for more details.
\begin{align}
\label{eq:control}
C(\hat Z_t,t,*) = \frac{\nabla\log P(\hat Z_t,t) - \nabla\log P(\Phi_t(Z_0),t)}{\sigma(t)}
\end{align}

\subsection{Spatiotemporal enhancement for CDFV}
\label{subsec: improvement}
\noindent\textbf{Implicit Cross-Attention Guidance}.
Although CDFV extracted from Video DiTs theoretically captures semantic differences between $P_0$ and $P_1$ with temporal coherence (Sec~\ref{subsec: CDFV}), empirical studies reveal persistent background leakage (Fig.~\ref{fig: framework}). 
We attribute this phenomenon to the score function $\nabla_X \log p_t(X; \theta)$, which is learned by the model and may not perfectly align with theoretical expectations. This discrepancy can introduce local distributional drift in unedited regions, and such shifts have the potential to cause noticeable alterations in the background of edited videos (see Fig.~\ref{fig: ablation1} for examples).
Segmentation masks play a crucial role in effective structure guidance, and cross-attention, as highlighted in \cite{cai2024freemask,qi2023fatezero,hertz2022prompt}, exhibit significant potential for shape editing tasks. This is attributed to their time-aware adaptability and target-following characteristics, which enhance the capability to maintain structural integrity and motion consistency over time.
Although most recent Video DiTs have moved from discrete cross-attention to Full Attention~\cite{yang2024cogvideox} for more accurate spatial-temporal learning, we introduce Implicit Cross-Attention derived from Full Attention. ICA still retains the essence of traditional cross-attention and guides shape editing effectively. Given text embeddings $\textbf{E} \in \mathbb{R}^{N \times d}$ and latent video tokens $\textbf{B} \in \mathbb{R}^{M \times d}$, Full Attention mechanism first concatenates them to form a larger matrix $\textbf{C} = [\textbf{E}; \textbf{B}] \in \mathbb{R}^{(N+M) \times d}$, each row of $\textbf{C}$ can be considered as both Query ($Q$), Key ($K$), and Value ($V$). The full attention map is computed as follows:
\begin{equation}
\mathcal{A} = \text{Softmax}\left(\frac{\textbf{C} \textbf{C}^\top}{\sqrt{d}}\right) = 
\begin{bmatrix} 
\mathcal{A}_{EE} & \mathcal{A}_{EB} \\ 
\mathcal{A}_{BE} & \mathcal{A}_{BB} 
\end{bmatrix} \in \mathbb{R}^{(N+M)\times(N+M)}
\label{eq:full_attn}
\end{equation}
We identify that the off-diagonal block $\mathcal{A}_{EB}$ or $\mathcal{A}_{BE}$ inherently encodes cross-modal interactions. Our \textit{Implicit Cross-Attention} extracts this block of different timesteps and binarizes it into $M_t$. We mask $\Delta v_{t}(Z_0,c_0,c_1)$ with $M_t$ to restrain the changes in the unedited region as Eq.~\ref{eq:mask}. $M_t$ can also be optionally combined with the popular SAM \cite{kirillov2023segment} masks using Boolean operations.

\begin{equation}
\label{eq:mask}
\Delta v_{t,{M_t}}(Z_0,c_0,c_1)=M_t\odot\left [v_{t,c_1}(\hat Z_t)- v_{t,c_0}(\Phi_t(Z_0)) \right ]
\end{equation}

\noindent\textbf{Target Embedding Reinforcement}.
We observe that in 3D Full-Attention, the effect of text embeddings diminishes as frame length increases. This phenomenon is particularly evident in global editing tasks such as stylization. We attribute this issue to the competition between fixed-length text tokens $\mathbf{E} \in \mathbb{R}^{N \times d}$ and an increasing number of spatiotemporal tokens $\mathbf{Z} \in \mathbb{R}^{F \times H \times W \times d}$. As the video duration grows, vectors associated with stylization embeddings become increasingly sparse across frames. This sparsity may further reduce the guidance fidelity of the text embeddings. To address these challenges, we propose  Embedding Reinforcement (ER) for prompt alignment: 
\begin{equation}
\mathbf{\tilde{E}}^{(k)} = \mathbf{E} + \gamma^{(k)} \odot \mathbf{E}
\end{equation}
where $k$ is used to locate the target embedding for editing, and its value is amplified by $\gamma+1$. Specifically, we set $\gamma=0.2$ for shape editing and $\gamma=5$ for stylization. By reinforcing the embeddings, the cross-attention map is reweighted to focus on regions more relevant to the editing target, enhancing editing precision.

\section{Experimental results}
\label{sec:exp}

  \vspace{-0.6em}
\begin{figure*}[htbp] 
  \centering 
  \includegraphics[width=1\linewidth]{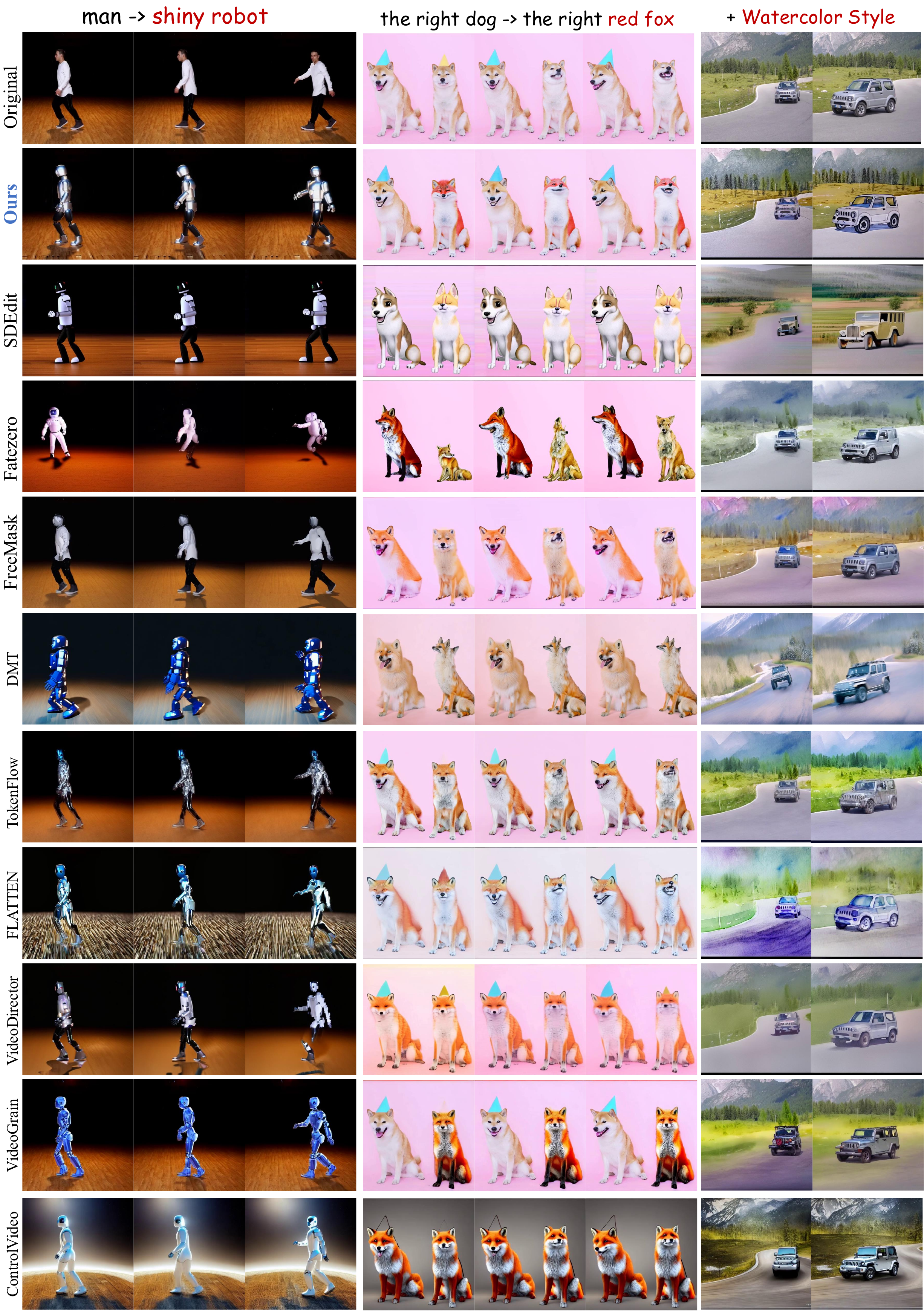} 
  \vspace{-1.6em}
  \caption{\textbf{Comparison.} Most methods based on attention-engineering and image diffusion models (FateZero~\cite{qi2023fatezero}, TokenFlow~\cite{geyer2023tokenflow}, VideoDirector~\cite{wang2024videodirector}) suffer from flickering and fail in multi-object editing. While VideoGrain~\cite{yangvideograin} enhances multi-object editing, it is inferior in structure consistency and motion detail fidelity (the second column). Attention-engineering-free approaches (FLATTEN~\cite{congflatten}, DMT~\cite{yatim2024space}, ControlVideo~\cite{zhang2023controlvideo}) exhibit structural infidelity. FreeMask~\cite{cai2024freemask} improves temporal consistency but remains constrained by its 3D-Unet base model. Applying the image editing method SDEdit~\cite{meng2021sdedit} directly to VideoDiTS compromises spatial-temporal fidelity. In comparison, our method achieves SOTA performance in fidelity, alignment, and temporal consistency. Refer to the supplementary material for more results.} 
  \label{fig:compare2} 
\end{figure*}
\textbf{Experimental setup.} We adopt the pretrained CovideoX-5B~\cite{yang2024cogvideox} as the base model and also extend our method to Wan2.1-14B~\cite{wang2025wan} to demonstrate the robustness and flexibility of DEVEdit. All experiments are conducted on one A100-80G GPU. We evaluate our methods on public DAVIS2017~\cite{pont20172017} videos and Internet open-source videos from Pexels~\cite{pexels}. In comparison experiments, we test 40-frame videos with a resolution of 512 $\times$ 512. Our focus is on training-free appearance editing, including local editing (shape and attribute editing), and global editing (stylization). 

\textbf{Baselines.} For baselines, we compare against image diffusion-based training-free editing methods, including FateZero~\cite{qi2023fatezero}, TokenFlow~\cite{geyer2023tokenflow}, VideoDirector~\cite{wang2024videodirector}, and VideoGain~\cite{yangvideograin}, which rely on attention engineering; ControlVideo~\cite{zhang2023controlvideo}, FLATTEN~\cite{congflatten}, and DMT~\cite{yatim2024space}, which are free of attention engineering; FreeMask~\cite{cai2024freemask}, which is based on a U-net-based video diffusion model with attention engineering; and SDEdit~\cite{meng2021sdedit} (directly applied to CogVideoX-5B~\cite{yang2024cogvideox} base model for video editing). 


\begin{table}[t]
  \centering
  \renewcommand{\arraystretch}{1.2} 
  \caption{\textbf{Quantitative evaluation and user study results.}}
  \resizebox{\linewidth}{!}{
  \begin{tabular}{@{} l c c c c c c c c c c c c @{}}
    \toprule
    
    \Large Method & \multicolumn{2}{c}{\Large Consistency} 
           & \multicolumn{2}{c}{\Large Fidelity} 
           & \Large Alignment 
           & \multicolumn{3}{c}{\Large User Study} 
           & \multicolumn{3}{c}{\Large Computation Efficiency}\\
    \cmidrule(lr){2-3} \cmidrule(lr){4-5} \cmidrule(lr){6-6}\cmidrule(lr){7-9} \cmidrule(lr){10-12}
    & \Large CLIP-F${\uparrow}$ & \Large E$_{warp} {\downarrow}$ 
    & \Large M.PSNR${\uparrow}$ & \Large LPIPS${\downarrow}$ 
    & \Large CLIP-T${\uparrow}$ 
    & \Large Edit${\uparrow}$ & \Large Quality${\uparrow}$ & \Large Consistency${\uparrow}$ &\Large VRAM ${\downarrow}$   &\Large RAM${\downarrow}$ &\Large Latency${\downarrow}$ \\
    
    \midrule
    \Large SDEdit~\cite{meng2021sdedit}                  & \Large \underline{0.9811} & \Large 1.67              & \Large 20.52              &\Large 0.4090              & \Large 27.46     &\Large 66.57  &\Large \underline{80.45}  &\Large \underline{85.66}  &\Large \underline{1.01}  &\Large \underline{1.13} &\Large \textbf{0.87}  \\
    \Large FateZero~\cite{qi2023fatezero}                & \Large 0.9289             & \Large 3.09              & \Large 23.39              &\Large 0.2634              & \Large 26.08     &\Large 58.87  &\Large 50.63  &\Large 56.89   &\Large 2.32 &\Large 21.44 &\Large 3.40 \\
    \Large FreeMask~\cite{cai2024freemask}               & \Large 0.9699             & \Large 2.00              & \Large 29.92              &\Large 0.2314              & \Large 27.06     &\Large 75.88  &\Large 74.67  &\Large 77.13   &\Large 1.64  &\Large 25.58 &\Large 5.65  \\
    \Large Tokenflow~\cite{geyer2023tokenflow}           & \Large 0.9583             & \Large \underline{1.48}  & \Large 29.97              &\Large \underline{0.2247}  & \Large \underline{29.78} &\Large 70.12  &\Large 53.45  &\Large 57.41   &\Large 1.43  &\Large 3.69 &\Large 13.03  \\
    \Large VideoDirector~\cite{wang2024videodirector}    & \Large 0.9555             & \Large 2.44              & \Large 28.97              &\Large 0.3205              & \Large 27.50     &\Large 74.13  &\Large 73.25  &\Large 71.45   &\Large 6.00  &\Large 2.26 &\Large 27.97 \\
    \Large VideoGrain~\cite{yangvideograin}              & \Large 0.9695             & \Large 2.68              & \Large \underline{30.70}  &\Large 0.2948              & \Large 27.79     &\Large \underline{76.41}  &\Large 79.87  &\Large 70.61   &\Large 2.35  &\Large 2.61 &\Large 13.44 \\
    \Large FLATTEN~\cite{congflatten}                    & \Large 0.9510             & \Large 4.89              & \Large 15.91              &\Large 0.3559              & \Large 27.57     &\Large 63.45  &\Large 69.45  &\Large 68.32   &\Large 1.54  &\Large 7.31 &\Large 4.61  \\
    \Large ControlVideo ~\cite{zhang2023controlvideo}    & \Large 0.9533             & \Large 3.10              & \Large 10.08              &\Large 0.4015              & \Large 27.06     &\Large 56.08  &\Large 55.33  &\Large 59.41   &\Large 8.74  &\Large 1.62 &\Large 9.45  \\
    \Large DMT~\cite{yatim2024space}                     & \Large 0.9668             & \Large 3.50              & \Large 15.95              &\Large 0.5096              & \Large 25.34     &\Large 62.66  &\Large 68.36  &\Large 69.88   &\Large 5.64  &\Large 3.32 &\Large 24.40 \\
    \rowcolor{gray!25}\Large DFVEdit                     & \Large \textbf{0.9924}    &\Large \textbf{1.12}      &\Large \textbf{31.18}      &\Large \textbf{0.1886}     & \Large \textbf{30.84}     &\Large \textbf{87.65}  &\Large \textbf{84.56} &\Large \textbf{86.98}    &\Large \textbf{0.95}  &\Large \textbf{0.86} &\Large \underline{1.20} \\
    \midrule
    \Large w/o ICA                                       & \Large 0.9922   &\Large 1.25   & \Large 29.33  &\Large 0.1920 &\Large 31.02      &\Large 86.45  &\Large 84.33 &\Large 86.56  &\Large 0.94 &\Large 0.78 &\Large 1.19 \\
    \Large w/o EmbedRF                                   & \Large 0.9913    &\Large 1.13   & \Large 31.15  &\Large 0.1889 &\Large 29.25      &\Large 86.04  &\Large 83.15  &\Large 86.13  &\Large 0.95 &\Large 0.85 &\Large 1.20 \\
    \bottomrule
  \end{tabular}
  }
  \label{table:quantitative_baseline}
\end{table}

\begin{figure*}[htbp] 
  \centering 
  \includegraphics[width=1\linewidth]{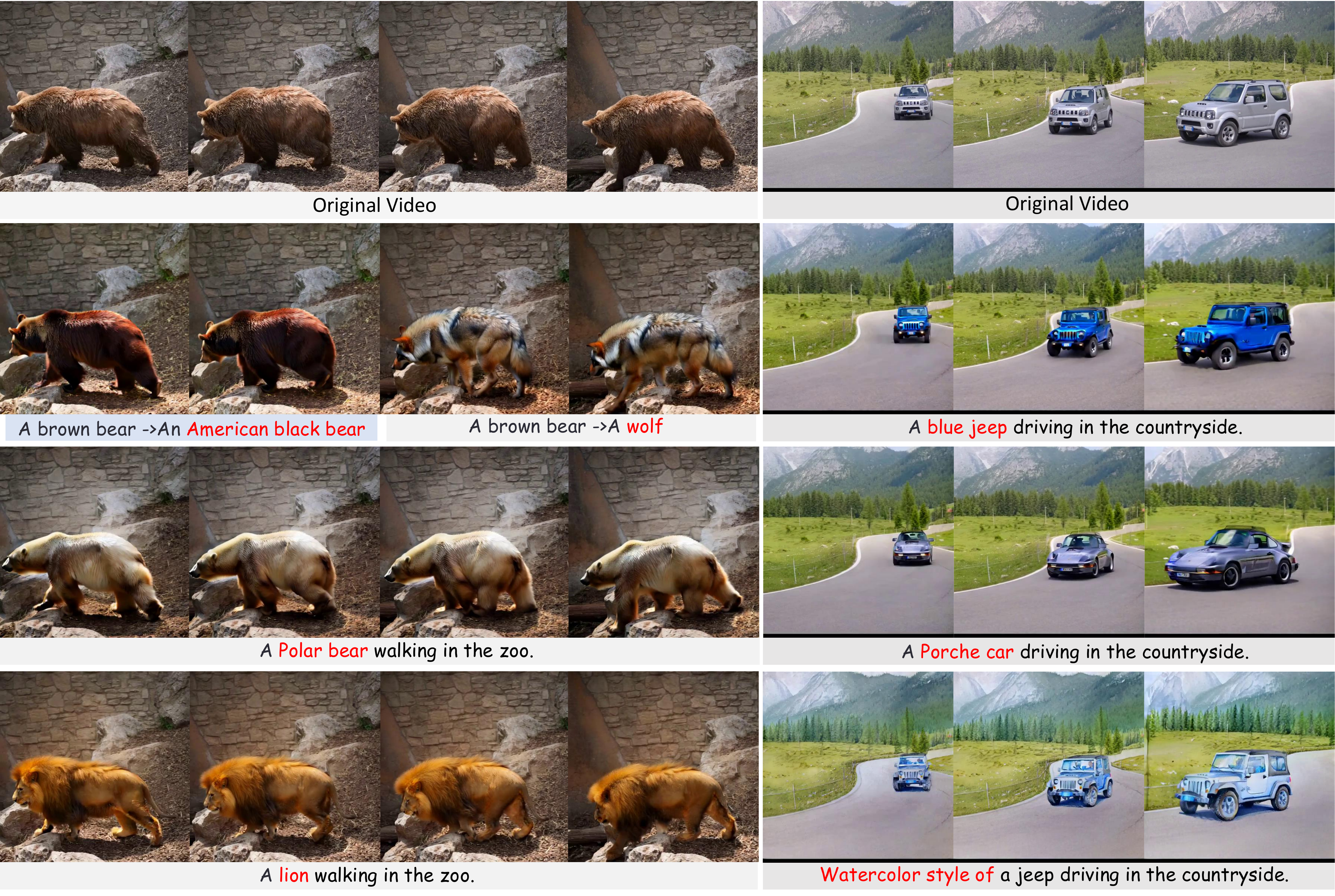} 
    \vspace{-1.5em}
    \caption{\textbf{Extensive qualitative results.} The extensive experiments take Wan2.1-14B~\cite{wang2025wan} as the base model, demonstrating the generalization of DFVEdit for Video DiTs. See the supplementary material for more results.}
  \label{fig:compare1} 
\end{figure*}

\subsection{Evaluation}

\textbf{Qualitative evaluation.} 
Fig.~\ref{fig:compare2} provides qualitative comparison results, showcasing our method's superiority in structure fidelity, motion integrity, and temporal consistency over other prominent baselines. For \textbf{single object editing} (first column), FateZero~\cite{qi2023fatezero}, TokenFlow~\cite{geyer2023tokenflow}, and VideoDirector~\cite{wang2024videodirector} exhibit noticeable flickering, while ControlVideo~\cite{zhang2023controlvideo}, FLATTEN~\cite{congflatten}, and DMT~\cite{yatim2024space} fail to preserve the details of unedited regions. For \textbf{multi-object editing} (second column), most methods struggle with editing accuracy; although VideoGrain~\cite{yangvideograin} achieves success in multi-object editing using fine-grained SAM~\cite{kirillov2023segment} masks, it falls short in maintaining motion detail fidelity (e.g., a mismatch between the fox and dog expressions). For \textbf{stylization} (third column), Freemask~\cite{cai2024freemask}, which is based on a UNet-based video diffusion model, performs notably well, while other methods still show inconsistencies in color tone and structural details (refer to the supplementary material for video displays). Additionally, we extended FateZero~\cite{qi2023fatezero} and KVEdit~\cite{zhu2025kv} directly to Cogvideo-5B~\cite{yang2024cogvideox} to compare editing quality and efficiency. Due to space limitations, please refer to the appendix for more detailed comparison results. Fig.~\ref{fig:compare1} provides the extensive experiment results on Wan2.1-14B~\cite{wang2025wan}, which also demonstrates high editing quality with respect to structure fidelity, motion integrity, and prompt alignment. Wan~\cite{wang2025wan} is combined with Flow Matching~\cite{lipman2022flow}, while Cogvideox~\cite{yang2024cogvideox} is based on Score Matching~\cite{song2020score}. As illustrated in both Fig.~\ref{fig:compare1} and Fig.~\ref{fig:compare2}, DFVEdit achieves consistent editing quality across popular Video DiTs, whether based on Score Matching~\cite{song2020score} or Flow Matching~\cite{lipman2022flow}.

\textbf{Quantitative evaluation.} In Tab.~\ref{table:quantitative_baseline}, we compare with baselines using both automatic metrics and human evaluations, following ~\cite{qi2023fatezero,liu2024video,cai2024freemask,geyer2023tokenflow,wu2023tune}. Specifically, \textbf{CLIP-F} calculates inter-frame cosine similarity to assess structural consistency, while \textbf{$\mathbf{E_{warp}}$} measures warping error~\cite{geyer2023tokenflow} to evaluate motion fidelity. Additionally, \textbf{M.PSNR} computes the Masked Peak Signal-to-Noise Ratio between source and target videos to gauge the fidelity of unedited regions, and \textbf{LPIPS} evaluates the Learned Perceptual Image Patch Similarity for overall structural fidelity. Moreover, \textbf{CLIP-T} quantifies the alignment between the target prompt and video through the CLIP Score~\cite{hessel2021clipscore}. Regarding user studies, we focus on Target Prompt Alignment (\textbf{Edit}), Overall Editing Quality including fidelity of unedited areas, minimal filtering and blurring (\textbf{Quality}), and Motion and Structural Consistency (\textbf{Consistency}). The results demonstrate that DFVEdit achieves superior spatial-temporal consistency, fidelity, and prompt alignment compared to other methods. Furthermore, to evaluate memory and computational efficiency, we measure Relative GPU Memory Consumption (\textbf{VRAM}), defined as the ratio of editing consumption on GPU relative to original inference consumption; Relative Inference Latency (\textbf{Latency}), which assesses the ratio of editing latency to inference latency; and Relative CPU Memory Consumption (\textbf{RAM}), measuring the ratio of editing consumption on CPU over original inference consumption. These metrics highlight the practical efficiency of DFVEdit. We also extend FateZero~\cite{qi2023fatezero} and KVEdit~\cite{zhu2025kv} to CogVideoX-5B~\cite{yang2024cogvideox} to evaluate their efficiencies. Some findings are illustrated in Fig.~\ref{fig:insight1}(b), demonstrating that these methods, originally designed for image diffusion with attention engineering, incur significant computational overhead when applied to Video DiTs.

\subsection{Ablation results}
\begin{figure*}[htbp] 
  \centering 
    \vspace{-0.5em}
  \includegraphics[width=1\textwidth]{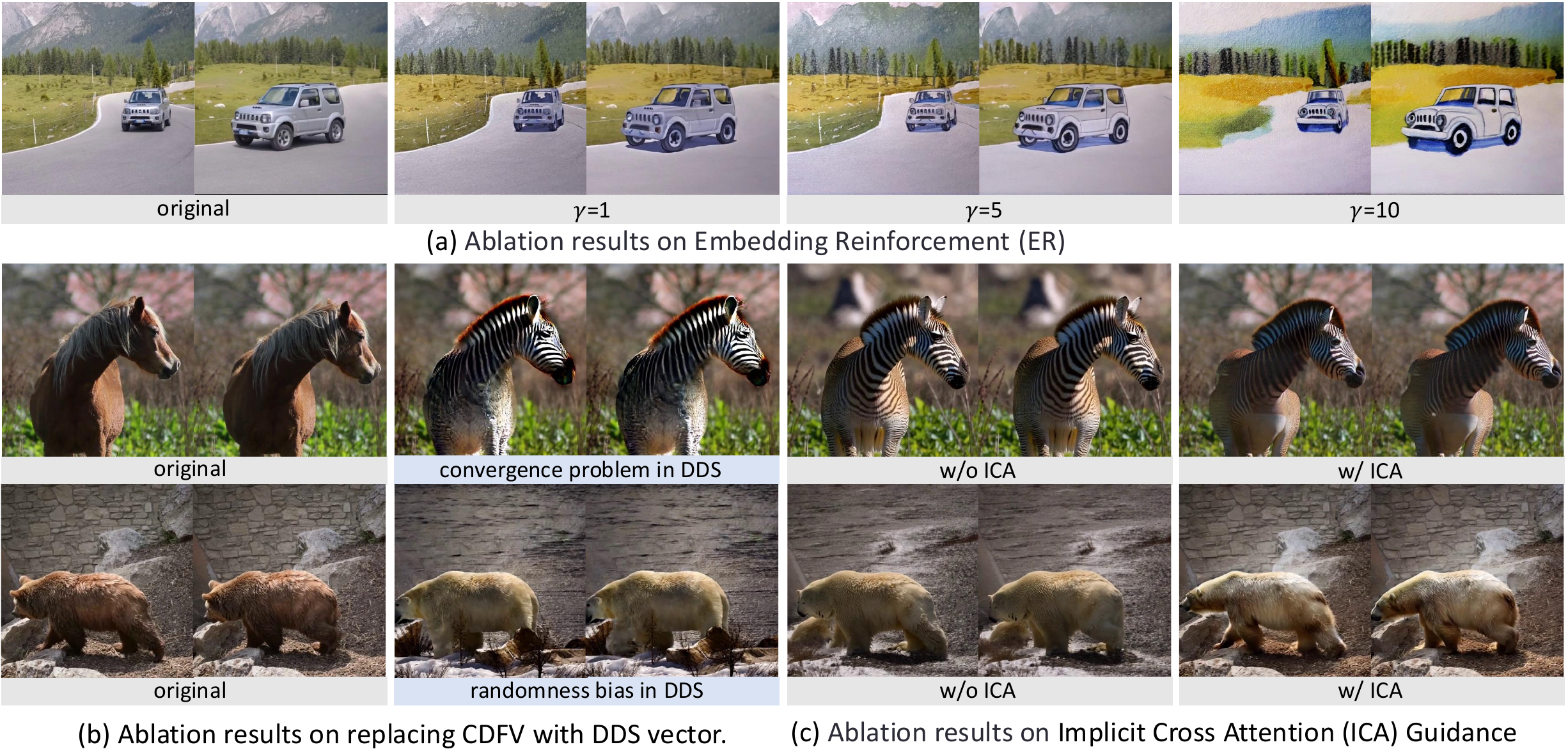} 
    \vspace{-1.5em}
  \caption{\textbf{Ablation.} (a)(c) demonstrate the effectiveness of ER and ICA. (b) highlights limitations of popular approximation-based latent refinement methods~\cite{hertz2023delta} in video editing, including: low convergence leading to unnatural changes and unpredictable convergence times; randomness bias resulting in unsatisfactory structural fidelity. Refer to the supplementary material for more results.} 
  \label{fig: ablation1} 
\end{figure*}

We evaluate the efficacy of CDFV, ICA, and ER in our ablation study. In Fig.~\ref{fig: ablation1}(a), we vary the Embedding Reinforcement factor $\gamma$ from 1 to 10. Without reinforcement ($\gamma=1$), stylization effects are negligible. Stylization improves as $\gamma$ increases but degrades with excessively high values. Empirically, $\gamma=5$ optimizes stylization without compromising structural fidelity or visual quality. Fig.~\ref{fig: ablation1}(c) shows that omitting Implicit Cross-Attention Guidance leads to unintended changes in unedited regions. Incorporating cross-attention mechanisms significantly enhances structural fidelity and overall quality. In Fig.~\ref{fig: ablation1}(b), we replace CDFV with the stochastic latent refinement vector in DDS~\cite{hertz2023delta}. In this ablation, for 'horse' experiment, ICA and ER are kept, while for 'bear' they are omitted for a fair comparison. The results highlights the effectiveness of CDFV. For additional qualitative and quantitative comparison and ablation results, please refer to the Appendix.

\vspace{-1em}
\section{Conclusion}
We present DFVEdit, an efficient and effective zero-shot video editing framework tailored for  Video Diffusion Transformers. DFVEdit realizes video editing through the direct flow transformation of the clean source latent. We theoretically unify editing and sampling from the continuous flow perspective, propose CDFV to estimate the flow vector from the source video to the target video, and further enhance the editing quality with ICA guidance and ER mechanism. Extensive experiments demonstrate the efficacy of DFVEdit on Video DiTs.

\bibliography{reference}

\begin{thebibliography}{10}

\bibitem{yang2024cogvideox}
Zhuoyi Yang, Jiayan Teng, Wendi Zheng, Ming Ding, Shiyu Huang, Jiazheng Xu, Yuanming Yang, Wenyi Hong, Xiaohan Zhang, Guanyu Feng, et~al.
\newblock Cogvideox: Text-to-video diffusion models with an expert transformer.
\newblock {\em arXiv preprint arXiv:2408.06072}, 2024.

\bibitem{kong2024hunyuanvideo}
Weijie Kong, Qi~Tian, Zijian Zhang, Rox Min, Zuozhuo Dai, Jin Zhou, Jiangfeng Xiong, Xin Li, Bo~Wu, Jianwei Zhang, et~al.
\newblock Hunyuanvideo: A systematic framework for large video generative models.
\newblock {\em arXiv preprint arXiv:2412.03603}, 2024.

\bibitem{peebles2023scalable}
William Peebles and Saining Xie.
\newblock Scalable diffusion models with transformers.
\newblock In {\em Proceedings of the IEEE/CVF International Conference on Computer Vision}, pages 4195--4205, 2023.

\bibitem{wang2025wan}
Ang Wang, Baole Ai, Bin Wen, Chaojie Mao, Chen-Wei Xie, Di~Chen, Feiwu Yu, Haiming Zhao, Jianxiao Yang, Jianyuan Zeng, et~al.
\newblock Wan: Open and advanced large-scale video generative models.
\newblock {\em arXiv preprint arXiv:2503.20314}, 2025.

\bibitem{feng2024dit4edit}
Kunyu Feng, Yue Ma, Bingyuan Wang, Chenyang Qi, Haozhe Chen, Qifeng Chen, and Zeyu Wang.
\newblock Dit4edit: Diffusion transformer for image editing.
\newblock {\em arXiv preprint arXiv:2411.03286}, 2024.

\bibitem{kulikov2024flowedit}
Vladimir Kulikov, Matan Kleiner, Inbar Huberman-Spiegelglas, and Tomer Michaeli.
\newblock Flowedit: Inversion-free text-based editing using pre-trained flow models.
\newblock {\em arXiv preprint arXiv:2412.08629}, 2024.

\bibitem{zhu2025kv}
Tianrui Zhu, Shiyi Zhang, Jiawei Shao, and Yansong Tang.
\newblock Kv-edit: Training-free image editing for precise background preservation.
\newblock {\em arXiv preprint arXiv:2502.17363}, 2025.

\bibitem{dalva2024fluxspace}
Yusuf Dalva, Kavana Venkatesh, and Pinar Yanardag.
\newblock Fluxspace: Disentangled semantic editing in rectified flow transformers.
\newblock {\em arXiv preprint arXiv:2412.09611}, 2024.

\bibitem{rout2024semantic}
Litu Rout, Yujia Chen, Nataniel Ruiz, Constantine Caramanis, Sanjay Shakkottai, and Wen-Sheng Chu.
\newblock Semantic image inversion and editing using rectified stochastic differential equations.
\newblock {\em arXiv preprint arXiv:2410.10792}, 2024.

\bibitem{jiao2025uniedit}
Guanlong Jiao, Biqing Huang, Kuan-Chieh Wang, and Renjie Liao.
\newblock Uniedit-flow: Unleashing inversion and editing in the era of flow models.
\newblock {\em arXiv preprint arXiv:2504.13109}, 2025.

\bibitem{singer2022make}
Uriel Singer, Adam Polyak, Thomas Hayes, Xi~Yin, Jie An, Songyang Zhang, Qiyuan Hu, Harry Yang, Oron Ashual, Oran Gafni, et~al.
\newblock Make-a-video: Text-to-video generation without text-video data.
\newblock {\em arXiv preprint arXiv:2209.14792}, 2022.

\bibitem{wu2023tune}
Jay~Zhangjie Wu, Yixiao Ge, Xintao Wang, Stan~Weixian Lei, Yuchao Gu, Yufei Shi, Wynne Hsu, Ying Shan, Xiaohu Qie, and Mike~Zheng Shou.
\newblock Tune-a-video: One-shot tuning of image diffusion models for text-to-video generation.
\newblock In {\em Proceedings of the IEEE/CVF International Conference on Computer Vision}, pages 7623--7633, 2023.

\bibitem{shin2024edit}
Chaehun Shin, Heeseung Kim, Che~Hyun Lee, Sang-gil Lee, and Sungroh Yoon.
\newblock Edit-a-video: Single video editing with object-aware consistency.
\newblock In {\em Asian Conference on Machine Learning}, pages 1215--1230. PMLR, 2024.

\bibitem{liu2024videop2p}
Shaoteng Liu, Yuechen Zhang, Wenbo Li, Zhe Lin, and Jiaya Jia.
\newblock Video-p2p: Video editing with cross-attention control.
\newblock In {\em Proceedings of the IEEE/CVF Conference on Computer Vision and Pattern Recognition}, pages 8599--8608, 2024.

\bibitem{qi2023fatezero}
Chenyang Qi, Xiaodong Cun, Yong Zhang, Chenyang Lei, Xintao Wang, Ying Shan, and Qifeng Chen.
\newblock Fatezero: Fusing attentions for zero-shot text-based video editing.
\newblock In {\em Proceedings of the IEEE/CVF International Conference on Computer Vision}, pages 15932--15942, 2023.

\bibitem{cai2024freemask}
Lingling Cai, Kang Zhao, Hangjie Yuan, Yingya Zhang, Shiwei Zhang, and Kejie Huang.
\newblock Freemask: Rethinking the importance of attention masks for zero-shot video editing.
\newblock {\em arXiv preprint arXiv:2409.20500}, 2024.

\bibitem{geyer2023tokenflow}
Michal Geyer, Omer Bar-Tal, Shai Bagon, and Tali Dekel.
\newblock Tokenflow: Consistent diffusion features for consistent video editing.
\newblock {\em arXiv preprint arXiv:2307.10373}, 2023.

\bibitem{zhang2023controlvideo}
Yabo Zhang, Yuxiang Wei, Dongsheng Jiang, Xiaopeng Zhang, Wangmeng Zuo, and Qi~Tian.
\newblock Controlvideo: Training-free controllable text-to-video generation.
\newblock {\em arXiv preprint arXiv:2305.13077}, 2023.

\bibitem{yangvideograin}
Xiangpeng Yang, Linchao Zhu, Hehe Fan, and Yi~Yang.
\newblock Videograin: Modulating space-time attention for multi-grained video editing.
\newblock In {\em The Thirteenth International Conference on Learning Representations}.

\bibitem{wang2024videodirector}
Yukun Wang, Longguang Wang, Zhiyuan Ma, Qibin Hu, Kai Xu, and Yulan Guo.
\newblock Videodirector: Precise video editing via text-to-video models.
\newblock {\em arXiv preprint arXiv:2411.17592}, 2024.

\bibitem{rombach2022high}
Robin Rombach, Andreas Blattmann, Dominik Lorenz, Patrick Esser, and Bj{\"o}rn Ommer.
\newblock High-resolution image synthesis with latent diffusion models.
\newblock In {\em Proceedings of the IEEE/CVF conference on computer vision and pattern recognition}, pages 10684--10695, 2022.

\bibitem{song2020denoising}
Jiaming Song, Chenlin Meng, and Stefano Ermon.
\newblock Denoising diffusion implicit models.
\newblock {\em arXiv preprint arXiv:2010.02502}, 2020.

\bibitem{khachatryan2023text2video}
Levon Khachatryan, Andranik Movsisyan, Vahram Tadevosyan, Roberto Henschel, Zhangyang Wang, Shant Navasardyan, and Humphrey Shi.
\newblock Text2video-zero: Text-to-image diffusion models are zero-shot video generators.
\newblock In {\em Proceedings of the IEEE/CVF International Conference on Computer Vision}, pages 15954--15964, 2023.

\bibitem{yatim2024space}
Danah Yatim, Rafail Fridman, Omer Bar-Tal, Yoni Kasten, and Tali Dekel.
\newblock Space-time diffusion features for zero-shot text-driven motion transfer.
\newblock In {\em 2024 IEEE/CVF Conference on Computer Vision and Pattern Recognition (CVPR)}, pages 8466--8476. IEEE Computer Society, 2024.

\bibitem{ku2024anyv2v}
Max Ku, Cong Wei, Weiming Ren, Huan Yang, and Wenhu Chen.
\newblock Anyv2v: A plug-and-play framework for any video-to-video editing tasks.
\newblock {\em arXiv e-prints}, pages arXiv--2403, 2024.

\bibitem{wang2023modelscope}
Jiuniu Wang, Hangjie Yuan, Dayou Chen, Yingya Zhang, Xiang Wang, and Shiwei Zhang.
\newblock Modelscope text-to-video technical report.
\newblock {\em arXiv preprint arXiv:2308.06571}, 2023.

\bibitem{li2024hunyuan}
Zhimin Li, Jianwei Zhang, Qin Lin, Jiangfeng Xiong, Yanxin Long, Xinchi Deng, Yingfang Zhang, Xingchao Liu, Minbin Huang, Zedong Xiao, et~al.
\newblock Hunyuan-dit: A powerful multi-resolution diffusion transformer with fine-grained chinese understanding.
\newblock {\em arXiv preprint arXiv:2405.08748}, 2024.

\bibitem{yang20241}
Chenglin Yang, Celong Liu, Xueqing Deng, Dongwon Kim, Xing Mei, Xiaohui Shen, and Liang-Chieh Chen.
\newblock 1.58-bit flux.
\newblock {\em arXiv preprint arXiv:2412.18653}, 2024.

\bibitem{song2020score}
Yang Song, Jascha Sohl-Dickstein, Diederik~P Kingma, Abhishek Kumar, Stefano Ermon, and Ben Poole.
\newblock Score-based generative modeling through stochastic differential equations.
\newblock {\em arXiv preprint arXiv:2011.13456}, 2020.

\bibitem{lipman2022flow}
Yaron Lipman, Ricky~TQ Chen, Heli Ben-Hamu, Maximilian Nickel, and Matt Le.
\newblock Flow matching for generative modeling.
\newblock {\em arXiv preprint arXiv:2210.02747}, 2022.

\bibitem{hertz2023delta}
Amir Hertz, Kfir Aberman, and Daniel Cohen-Or.
\newblock Delta denoising score.
\newblock In {\em Proceedings of the IEEE/CVF International Conference on Computer Vision}, pages 2328--2337, 2023.

\bibitem{pooledreamfusion}
Ben Poole, Ajay Jain, Jonathan~T Barron, and Ben Mildenhall.
\newblock Dreamfusion: Text-to-3d using 2d diffusion.
\newblock In {\em The Eleventh International Conference on Learning Representations}.

\bibitem{zhang2023i2vgen}
Shiwei Zhang, Jiayu Wang, Yingya Zhang, Kang Zhao, Hangjie Yuan, Zhiwu Qin, Xiang Wang, Deli Zhao, and Jingren Zhou.
\newblock I2vgen-xl: High-quality image-to-video synthesis via cascaded diffusion models.
\newblock {\em arXiv preprint arXiv:2311.04145}, 2023.

\bibitem{blattmann2023stable}
Andreas Blattmann, Tim Dockhorn, Sumith Kulal, Daniel Mendelevitch, Maciej Kilian, Dominik Lorenz, Yam Levi, Zion English, Vikram Voleti, Adam Letts, et~al.
\newblock Stable video diffusion: Scaling latent video diffusion models to large datasets.
\newblock {\em arXiv preprint arXiv:2311.15127}, 2023.

\bibitem{chen2024videocrafter2}
Haoxin Chen, Yong Zhang, Xiaodong Cun, Menghan Xia, Xintao Wang, Chao Weng, and Ying Shan.
\newblock Videocrafter2: Overcoming data limitations for high-quality video diffusion models.
\newblock In {\em Proceedings of the IEEE/CVF Conference on Computer Vision and Pattern Recognition}, pages 7310--7320, 2024.

\bibitem{zheng2024open}
Zangwei Zheng, Xiangyu Peng, Tianji Yang, Chenhui Shen, Shenggui Li, Hongxin Liu, Yukun Zhou, Tianyi Li, and Yang You.
\newblock Open-sora: Democratizing efficient video production for all.
\newblock {\em arXiv preprint arXiv:2412.20404}, 2024.

\bibitem{lin2024open}
Bin Lin, Yunyang Ge, Xinhua Cheng, Zongjian Li, Bin Zhu, Shaodong Wang, Xianyi He, Yang Ye, Shenghai Yuan, Liuhan Chen, et~al.
\newblock Open-sora plan: Open-source large video generation model.
\newblock {\em arXiv preprint arXiv:2412.00131}, 2024.

\bibitem{lu2022dpm}
Cheng Lu, Yuhao Zhou, Fan Bao, Jianfei Chen, Chongxuan Li, and Jun Zhu.
\newblock Dpm-solver++: Fast solver for guided sampling of diffusion probabilistic models.
\newblock {\em arXiv preprint arXiv:2211.01095}, 2022.

\bibitem{garibi2024renoise}
Daniel Garibi, Or~Patashnik, Andrey Voynov, Hadar Averbuch-Elor, and Daniel Cohen-Or.
\newblock Renoise: Real image inversion through iterative noising.
\newblock In {\em European Conference on Computer Vision}, pages 395--413. Springer, 2024.

\bibitem{deutch2024turboedit}
Gilad Deutch, Rinon Gal, Daniel Garibi, Or~Patashnik, and Daniel Cohen-Or.
\newblock Turboedit: Text-based image editing using few-step diffusion models.
\newblock In {\em SIGGRAPH Asia 2024 Conference Papers}, pages 1--12, 2024.

\bibitem{esser2024scaling}
Patrick Esser, Sumith Kulal, Andreas Blattmann, Rahim Entezari, Jonas M{\"u}ller, Harry Saini, Yam Levi, Dominik Lorenz, Axel Sauer, Frederic Boesel, et~al.
\newblock Scaling rectified flow transformers for high-resolution image synthesis.
\newblock In {\em Forty-first international conference on machine learning}, 2024.

\bibitem{sauer2024adversarial}
Axel Sauer, Dominik Lorenz, Andreas Blattmann, and Robin Rombach.
\newblock Adversarial diffusion distillation.
\newblock In {\em European Conference on Computer Vision}, pages 87--103. Springer, 2024.

\bibitem{nguyen2024swiftedit}
Trong-Tung Nguyen, Quang Nguyen, Khoi Nguyen, Anh Tran, and Cuong Pham.
\newblock Swiftedit: Lightning fast text-guided image editing via one-step diffusion.
\newblock {\em arXiv preprint arXiv:2412.04301}, 2024.

\bibitem{jiang2025vace}
Zeyinzi Jiang, Zhen Han, Chaojie Mao, Jingfeng Zhang, Yulin Pan, and Yu~Liu.
\newblock Vace: All-in-one video creation and editing.
\newblock {\em arXiv preprint arXiv:2503.07598}, 2025.

\bibitem{peruzzo2024vase}
Elia Peruzzo, Vidit Goel, Dejia Xu, Xingqian Xu, Yifan Jiang, Zhangyang Wang, Humphrey Shi, and Nicu Sebe.
\newblock Vase: Object-centric appearance and shape manipulation of real videos.
\newblock {\em arXiv preprint arXiv:2401.02473}, 2024.

\bibitem{esser2023structure}
Patrick Esser, Johnathan Chiu, Parmida Atighehchian, Jonathan Granskog, and Anastasis Germanidis.
\newblock Structure and content-guided video synthesis with diffusion models.
\newblock In {\em Proceedings of the IEEE/CVF international conference on computer vision}, pages 7346--7356, 2023.

\bibitem{gu2024videoswap}
Yuchao Gu, Yipin Zhou, Bichen Wu, Licheng Yu, Jia-Wei Liu, Rui Zhao, Jay~Zhangjie Wu, David~Junhao Zhang, Mike~Zheng Shou, and Kevin Tang.
\newblock Videoswap: Customized video subject swapping with interactive semantic point correspondence.
\newblock In {\em Proceedings of the IEEE/CVF Conference on Computer Vision and Pattern Recognition}, pages 7621--7630, 2024.

\bibitem{zi2025cococo}
Bojia Zi, Shihao Zhao, Xianbiao Qi, Jianan Wang, Yukai Shi, Qianyu Chen, Bin Liang, Rong Xiao, Kam-Fai Wong, and Lei Zhang.
\newblock Cococo: Improving text-guided video inpainting for better consistency, controllability and compatibility.
\newblock In {\em Proceedings of the AAAI Conference on Artificial Intelligence}, volume~39, pages 11067--11076, 2025.

\bibitem{wang2024videocomposer}
Xiang Wang, Hangjie Yuan, Shiwei Zhang, Dayou Chen, Jiuniu Wang, Yingya Zhang, Yujun Shen, Deli Zhao, and Jingren Zhou.
\newblock Videocomposer: Compositional video synthesis with motion controllability.
\newblock {\em Advances in Neural Information Processing Systems}, 36, 2024.

\bibitem{meng2021sdedit}
Chenlin Meng, Yutong He, Yang Song, Jiaming Song, Jiajun Wu, Jun-Yan Zhu, and Stefano Ermon.
\newblock Sdedit: Guided image synthesis and editing with stochastic differential equations.
\newblock {\em arXiv preprint arXiv:2108.01073}, 2021.

\bibitem{yang2023rerender}
Shuai Yang, Yifan Zhou, Ziwei Liu, and Chen~Change Loy.
\newblock Rerender a video: Zero-shot text-guided video-to-video translation.
\newblock In {\em SIGGRAPH Asia 2023 Conference Papers}, pages 1--11, 2023.

\bibitem{kirillov2023segment}
Alexander Kirillov, Eric Mintun, Nikhila Ravi, Hanzi Mao, Chloe Rolland, Laura Gustafson, Tete Xiao, Spencer Whitehead, Alexander~C Berg, Wan-Yen Lo, et~al.
\newblock Segment anything.
\newblock In {\em Proceedings of the IEEE/CVF international conference on computer vision}, pages 4015--4026, 2023.

\bibitem{vaswani2017attention}
Ashish Vaswani, Noam Shazeer, Niki Parmar, Jakob Uszkoreit, Llion Jones, Aidan~N Gomez, {\L}ukasz Kaiser, and Illia Polosukhin.
\newblock Attention is all you need.
\newblock {\em Advances in neural information processing systems}, 30, 2017.

\bibitem{yoon2024frag}
Sunjae Yoon, Gwanhyeong Koo, Geonwoo Kim, and Chang~D Yoo.
\newblock Frag: Frequency adapting group for diffusion video editing.
\newblock {\em arXiv preprint arXiv:2406.06044}, 2024.

\bibitem{liu2024stablev2v}
Chang Liu, Rui Li, Kaidong Zhang, Yunwei Lan, and Dong Liu.
\newblock Stablev2v: Stablizing shape consistency in video-to-video editing.
\newblock {\em arXiv preprint arXiv:2411.11045}, 2024.

\bibitem{kuanyv2v}
Max Ku, Cong Wei, Weiming Ren, Huan Yang, and Wenhu Chen.
\newblock Anyv2v: A tuning-free framework for any video-to-video editing tasks.
\newblock {\em Transactions on Machine Learning Research}.

\bibitem{songdenoising}
Jiaming Song, Chenlin Meng, and Stefano Ermon.
\newblock Denoising diffusion implicit models.
\newblock In {\em International Conference on Learning Representations}.

\bibitem{song2020improved}
Yang Song and Stefano Ermon.
\newblock Improved techniques for training score-based generative models.
\newblock {\em Advances in neural information processing systems}, 33:12438--12448, 2020.

\bibitem{song2019generative}
Yang Song and Stefano Ermon.
\newblock Generative modeling by estimating gradients of the data distribution.
\newblock {\em Advances in neural information processing systems}, 32, 2019.

\bibitem{uhlenbeck1930theory}
On the theory of the brownian motion.
\newblock {\em Physical review}, 36(5):823, 1930.

\bibitem{jordan1998variational}
Richard Jordan, David Kinderlehrer, and Felix Otto.
\newblock The variational formulation of the fokker--planck equation.
\newblock {\em SIAM journal on mathematical analysis}, 29(1):1--17, 1998.

\bibitem{kloeden1992stochastic}
Peter~E Kloeden, Eckhard Platen, Peter~E Kloeden, and Eckhard Platen.
\newblock {\em Stochastic differential equations}.
\newblock Springer, 1992.

\bibitem{ho2020denoising}
Jonathan Ho, Ajay Jain, and Pieter Abbeel.
\newblock Denoising diffusion probabilistic models.
\newblock {\em Advances in neural information processing systems}, 33:6840--6851, 2020.

\bibitem{han2024proxedit}
Ligong Han, Song Wen, Qi~Chen, Zhixing Zhang, Kunpeng Song, Mengwei Ren, Ruijiang Gao, Anastasis Stathopoulos, Xiaoxiao He, Yuxiao Chen, et~al.
\newblock Proxedit: Improving tuning-free real image editing with proximal guidance.
\newblock In {\em Proceedings of the IEEE/CVF Winter Conference on Applications of Computer Vision}, pages 4291--4301, 2024.

\bibitem{couairon2022diffedit}
Guillaume Couairon, Jakob Verbeek, Holger Schwenk, and Matthieu Cord.
\newblock Diffedit: Diffusion-based semantic image editing with mask guidance.
\newblock {\em arXiv preprint arXiv:2210.11427}, 2022.

\bibitem{hertz2022prompt}
Amir Hertz, Ron Mokady, Jay Tenenbaum, Kfir Aberman, Yael Pritch, and Daniel Cohen-Or.
\newblock Prompt-to-prompt image editing with cross attention control.
\newblock {\em arXiv preprint arXiv:2208.01626}, 2022.

\bibitem{congflatten}
Yuren Cong, Mengmeng Xu, Shoufa Chen, Jiawei Ren, Yanping Xie, Juan-Manuel Perez-Rua, Bodo Rosenhahn, Tao Xiang, Sen He, et~al.
\newblock Flatten: optical flow-guided attention for consistent text-to-video editing.
\newblock In {\em The Twelfth International Conference on Learning Representations}.

\bibitem{pont20172017}
Jordi Pont-Tuset, Federico Perazzi, Sergi Caelles, Pablo Arbel{\'a}ez, Alex Sorkine-Hornung, and Luc Van~Gool.
\newblock The 2017 davis challenge on video object segmentation.
\newblock {\em arXiv preprint arXiv:1704.00675}, 2017.

\bibitem{pexels}
Pexels.
\newblock Pexels free stock video clips and motion graphics.
\newblock \url{https://www.pexels.com}.
\newblock Accessed: 2025-05-15.

\bibitem{liu2024video}
Shaoteng Liu, Yuechen Zhang, Wenbo Li, Zhe Lin, and Jiaya Jia.
\newblock Video-p2p: Video editing with cross-attention control.
\newblock In {\em Proceedings of the IEEE/CVF Conference on Computer Vision and Pattern Recognition}, pages 8599--8608, 2024.

\bibitem{hessel2021clipscore}
Jack Hessel, Ari Holtzman, Maxwell Forbes, Ronan~Le Bras, and Yejin Choi.
\newblock Clipscore: A reference-free evaluation metric for image captioning.
\newblock {\em arXiv preprint arXiv:2104.08718}, 2021.

\end{thebibliography}

\newpage
\renewcommand{\thetable}{T\arabic{table}}  
\renewcommand{\thefigure}{F\arabic{figure}} 
\appendix

\section{Additional theoretical details}
\subsection{Revisiting video editing from sampling perspective}
Let $\{X_t^{\text{edit}}\}_{t=0}^T$ define the state trajectory of the edited video in the sampling process. We formalize video editing as a \textit{controlled Markov chain} with the following recursive relation:
\begin{equation}
    X_{t-1}^{\text{edit}} = g_{\theta_{2,t}}\Big( X_t^{\text{edit}},\ \underbrace{\epsilon_{\theta_1}(X_t^{\text{edit}}, t)}_{\text{Canonical Denoiser}} + \lambda \underbrace{C( X_t^{\text{edit}}, t,*)}_{\text{Control Term}} \Big)
\label{eq: edit}
\end{equation}
where \textit{State Transition} $g_{\theta_{2,t}}$ is the differentiable transition function parameterized by learnable $\theta_2$, $\epsilon_{\theta_1}$ is the pretrained diffusion model with frozen $\theta_1$, \textit{Control Term} $C$ is the editing condition injector with intensity $\lambda \geq 0$.

The formulation maintains \textbf{\textit{consistency with standard diffusion sampling process}} when $\lambda=0$ and $g_{\theta_2} = \mathcal{I}$, where $\mathcal{I}: \mathcal{X} \to \mathcal{X}$ denotes the identity operator satisfying $\mathcal{I}(x) = x,\ \forall x \in \mathcal{X}$.:

\begin{equation}
    X_{t-1}^{\text{edit}} \big|_{\substack{
        \lambda=0 \\ 
        g_{\theta_2} = \mathcal{I}
    }} \equiv X_{t-1}^{\text{orig}}
\end{equation}

\subsubsection{Unification with various editing methods}
Existing popular editing paradigms emerge as special cases of our control framework:

\begin{enumerate}[leftmargin=*]
    \item \textbf{Inversion-based editing} (like Fatezero~\cite{qi2023fatezero}):

    \begin{align}
        g_{\theta_{2,t}}(a,b) &= \frac{\sqrt{\alpha_{t-1}}}{\sqrt{\alpha_t}}\left(a + \Delta\beta_t b\right) \\
        C(X_t^{\text{edit}},t,*)&= \epsilon_{\theta_1}^{\text{edit}}(X_t^{\text{edit}},t) - \epsilon_{\theta_1}(X_t^{\text{edit}},t)\\
        \Delta\beta_t &= \sqrt{\frac{1-\alpha_{t-1}}{\alpha_{t-1}}} - \sqrt{\frac{1-\alpha_t}{\alpha_t}}
    \end{align}

    \item \textbf{Latent-approximation-based editing} (like DDS~\cite{hertz2023delta}):
    \begin{align}
        g_{\theta_{2,t}}(a,b) &= \text{Proj}_{\theta_{2,t}}(a + \eta b) \\
        C(x_t,t,*) &= \epsilon_{\theta_1}(x_t,t) - \epsilon_{\theta_1}(x_t,t) - \epsilon \\
        \epsilon &\sim \mathcal{N}(0, \sigma_t^2 I)
    \end{align}
\end{enumerate}

where $\alpha_t$ is the DDPM noise schedule coefficient at step $t$, $\Delta\beta_t$ is the noise scale difference term maintaining consistency in the reverse process, $\epsilon_{\theta_1}^{\text{edit}}$ is the edited noise prediction conditioned on the target prompt, $X_t^{\text{edit}}$ is the latent representation during the editing process. And ${Proj}_{\theta_{2,t}}$ is a shallow approximation network with learnable parameter $\theta_2$ that directly refines the latent to the target latent, $\eta$ is the step size controlling parameter update strength, $\sigma_t$ is the time-dependent noise scale for stochastic refinement, and $\epsilon$ is the Gaussian noise enabling exploration in the latent space. These formulations show how various editing methods are implicitly isomorphic with the sampling process.

\subsubsection{Revisiting video editing from the continuous flow transformation perspective} 
For the sampling process of SDE: 
\begin{align}
dX = \underbrace{-\frac{1}{2} \beta(t) X}_{f(X,t)} dt + \underbrace{\sqrt{\beta(t)}}_{g(t)} dW
\end{align}

For the inverse process of SDE: 
\begin{align}
dX = \left[ -\frac{1}{2} \beta(t) X - \beta(t) \nabla_X \log p_t(X) \right] dt + \sqrt{\beta(t)} dW
\end{align}

when changing the discrete update formulation into continuous $\Delta t \to 0$, we define:

\begin{align}
    \alpha(t) = e^{-\int_0^t \beta(s) ds}, \quad \sigma(t) = \sqrt{\frac{1-\alpha(t)}{\alpha(t)}}
\end{align}

Using Taylor's expansion,we have:

\begin{align}
\alpha_{t-1} \approx \alpha(t) - \dot{\alpha}(t) \Delta t \\
\frac{\sqrt{\alpha_{t-1}}}{\sqrt{\alpha_t}} \approx 1 - \frac{1}{2} \beta(t) \Delta t
\end{align}

\begin{align}
X_{t-1}^{\text{edit}} \approx X_t^{\text{edit}} - \frac{\beta(t)}{2} \left( X_t^{\text{edit}} + \sigma(t) (\epsilon_{\theta_3} + \lambda C) \right) \Delta t
\end{align}

Under the Euler discretization scheme with step size $\Delta t \to 0$ and $g_{\theta_2} = \mathcal{I}$, the discrete process \eqref{eq: edit} converges to the controlled SDE:

\begin{align}
    dX^{\text{edit}}_t = \underbrace{
  \left[ 
    -\frac{\beta(t)}{2} X^{\text{edit}}_t  + \frac{\beta(t)}{2} \nabla\log p_t(X^{\text{edit}}_t ) + \lambda \frac{\beta(t)}{2} \sigma(t) C(X^{\text{edit}}_t ,t)
  \right]
}_{f_{\theta_1}(X^{\text{edit}}_t ,t)} dt + \underbrace{\sqrt{\beta(t)}}_{g(t)} dW
\label{eq: edit2}
\end{align}

And our derived CDFV adheres to the \textit{minimum intervention principle} from optimal control theory, which theoretically guarantees computational efficiency:
\begin{align}
    \min_{\lambda, C} \mathbb{E} \left[ \int_0^T \| C(X_t,t) \|^2 dt \right] \quad \text{s.t.} \quad dX = \left[ f_{\theta_1}(X,t) + \lambda C(X,t) \right] dt + g(t) dW
\end{align}

\begin{align}
    C^*(X,t) = \frac{\nabla_X \log p_t^{\text{edit}}(X) - \nabla_X \log p_t(X)}{\sigma(t)}
\end{align}

In addition, we provide the simplified algorithm of DFVEdit as below:

\begin{algorithm}
\caption{Simplified algorithm for DFVEdit}
\label{alg:dfvedit_detail}
\begin{algorithmic}[1]
\Require 
    source video $\mathbf{X}_0$, 
    target and source prompt embeddings $[C_1, C_0]$,
    Video DiT $\epsilon_{\theta_1}$,
    encoder $\mathcal{E}(\cdot)$, decoder $\mathcal{D}(\cdot)$,
    sampling timesteps $T$,
    ER scale $\gamma^{(k)}$
\Ensure Edited video $\mathbf{X}_1$

\State $\mathbf{Z}_0 \leftarrow \mathcal{E}(\mathbf{X}_0)$ \Comment{Latent encoding}
\State $\hat{\mathbf{Z}}_T \leftarrow \mathbf{Z}_0$ \Comment{Initialize target latent}
\State $\tilde{C}_1 \leftarrow C_1 + \gamma^{(k)} \odot C_1$ 
    \Comment{Embedding Reinforcement}

\For{$t \leftarrow T$ \textbf{down to} $1$}
    \State $\mathbf{Z}_{\text{trans}} \leftarrow \Phi_t([\hat{\mathbf{Z}}_t; \mathbf{Z}_0])$ 
        \Comment{One-step forward process: $q(\mathbf{z}_t|\mathbf{z}_{0})$}
    
    \State $\Delta v_t \xleftarrow[\Delta ]{v_{(t.c_1)}\text{,}v_{(t,c_0)}} \epsilon_{\theta_1}(\mathbf{Z}_{\text{trans}}, [\tilde{C}_1,C_0])$ 
        \Comment{Raw CDFV prediction}
    
    \State $\Delta v_{(t,M_t)} \leftarrow M_t \odot \left[\Delta v_t\right]$ 
        \Comment{Implicit Cross-Attention Guidance}
    
    \State $\hat{\mathbf{Z}}_{t-1} \leftarrow \hat{\mathbf{Z}}_t - \Delta v_{(t,M_t)}$ 
        \Comment{Latent update}
\EndFor

\State $\mathbf{X}_1 \leftarrow \mathcal{D}(\hat{\mathbf{Z}}_0)$ \Comment{Video synthesis}
\Statex \hrulefill 
\Statex \textbf{Note}: 
\Statex \hspace{\algorithmicindent} $\triangleright$ Flow map $\Phi_t$ implements the one-step forward process with standard method-specific coefficients (DDPM/DDIM/Flow Matching). 
\Statex \hspace{\algorithmicindent} $\triangleright$ ICA mask $M_t$ is computed from the specific layer of the Full Attention map (Section 3.3).
\end{algorithmic}
\end{algorithm}

\section{Additional experimental details }

\subsection{Experimental settings}

We do quantitative and human evaluations with 8 quantitative metrics, including Temporal Consistency ('CLIP-F'), Warping Error ('$E_{warp}$')~\cite{geyer2023tokenflow}, Prompt Alignment ('CLIP-T'), Masked PSNR ('M.PSNR'), Perceptual Similarity ('LPIPS'), Relative GPU Memory Consumption ('VRAM'), Relative CPU Memory Consumption (‘RAM’), Relative Inference Latency ('Latency') and 3 metrics for user study, including Text Alignment ('Edit'), Overall Frame Quality ('Quality') and Temporal Consistency and Realism ('Consistency'). 

\noindent\textbf{CLIP metrics.} We employ the output logits of the official ViT-L-14 CLIP model to compute two metrics: (1) the mean cosine similarity between all frame embeddings and the target text prompt (CLIP-T), and (2) the average cosine similarity of consecutive frame embeddings of edited videos (CLIP-F).

\noindent\textbf{Masked PSNR.} We quantify structural preservation by computing Masked PSNR on unedited regions, following~\cite{liu2024video}. Using 10 DAVIS~\cite{pont20172017} videos, 30 diverse prompts and corresponding segmentation annotations $M$ provided by DAVIS, we calculate pixel-level differences between source ($\mathbf{X}_0$) and edited ($\mathbf{X}_1$) videos within regions identified by inverted segmentation masks $M^*=\neg M$. 
\begin{equation}
    M.PSNR(\mathbf{X_1},\mathbf{X_0}) =PSNR(B(\mathbf{X_1},M^*), B(\mathbf{X_1},M^*)
    \label{eq:3-3}
\end{equation}
where $B(\dots)$ is the binary operation with a threshold of 0.3. 

\noindent\textbf{User study.} We conducted a pairwise comparison study with 20 participants evaluating 80 video-prompt pairs (30 from DAVIS, 50 from the website Pexels~\cite{pexels}). Participants rated three aspects: (1) \textit{Text Alignment} (prompt-video correspondence), (2) \textit{Frame Quality} (visual artifacts), and (3) \textit{Consistency} (temporal coherence and motion preservation). Scores (0-100 scale) were aggregated by trimming extremes and averaging remaining responses, yielding 1600 total ratings.

\noindent\textbf{Relative Efficiency Metrics.} We evaluate computational efficiency through three normalized metrics: Relative CPU Memory Consumption means the ratio of average CPU Memory allocated during editing to that average CPU allocated to original inference (only generation) with the same base model. Relative Inference Latency means the ratio of the latency with editing to that of the original inference latency with the same model. These relative metrics enable fair comparison across varying base model requirements. Tab.~\ref{tab:app1} reports absolute values and experimental configurations. Reported values include absolute peak allocated GPU/CPU memory consumption (MB), processing latency, frame count ($F$), and corresponding base models. The groups 'Stable Diffusion', 'Zeroscope', and 'CogvideoX' represent the original generation results with base models, while other groups are the editing results.

\begin{table}[ht]
    \centering
    \caption{\textbf{Absolute empirical computational efficiency results.} }
    \label{tab:app1}
    \resizebox{\linewidth}{!}{\begin{tabular}{lrrrrl}
    \toprule
        Method & GPU Memory (MB) & CPU Memory (MB) & Latency (s) & $F$ & Base Model \\ \hline
        Stable Diffusion~\cite{rombach2022high} & 4134.31  & 2865.00  & 15.47  & 8 & Stable Diffusion 1.5 \\ 
        Zeroscope~\cite{wang2023modelscope} & 4551.32  & 3833.86  & 13.14  & 8 & Zeroscope \\ 
        CogVideoX~\cite{yang2024cogvideox} & 33110.36  & 10522.26  & 100.80  & 41 & CogVideoX-5B \\ 
        SDEdit~\cite{meng2021sdedit} & 33441.46 & 11890.15 & 87.69 & 41 & CogVideoX-5B \\ 
        FateZero~\cite{qi2023fatezero} & 9576.13  & 61416.08  & 52.58  & 8 & Stable Diffusion 1.5 \\ 
        FreeMask~\cite{cai2024freemask} & 7464.16  & 98059.40  & 74.29  & 8 & Zeroscope \\
        TokenFlow~\cite{geyer2023tokenflow} & 17040.29  & 159475.48  & 126.87  & 8 & Stable Diffusion 1.5 \\ 
        VideoDirector~\cite{wang2024videodirector} & 24789.38  & 6475.66  & 432.64  & 8 & Stable Diffusion 1.5 \\ 
        FLATTEN~\cite{congflatten} & 6385.01  & 20936.98  & 71.35  & 8 & Stable Diffusion 1.5 \\
        ControlVideo~\cite{zhang2023controlvideo} & 36115.23  & 4635.36  & 146.19  & 8 & Stable Diffusion 1.5 \\ 
        DMT~\cite{yatim2024space} & 42500.24  & 25572.34  & 217.54  & 8 & Stable Diffusion 1.5 \\ 
        Ours & 31454.84  & 9049.14  & 120.96  & 41 & CogVideoX-5B \\ \hline
    \end{tabular}}
\end{table}

\subsection{Experimental details on insight}
In Fig.~1 of the main text, we present some visualizations of our insights: (a) presents the theoretical attention memory consumption of different base models; (b) presents the theoretical inference latency of FateZero~\cite{qi2023fatezero} and KVEdit~\cite{zhu2025kv} when directly applying them to CogVideoX-5B, as well as the practical inference latency of DFVEdit on CogVideoX-5B. Here, we provide more details on these insights.

\noindent{\textbf{Attention memory explosion in DiT models.}} We analyze the memory consumption of attention mechanisms in the Unet module of diffusion models versus the Transformer module of DiT models, focusing on estimating the memory (GB) needed for storing attention score maps in float32 format per timestep. Although attention score maps are rarely computed explicitly in base models due to efficiency concerns, traditional editing methods often require their direct manipulation for attention engineering. Therefore, explicit examination of these maps helps in identifying challenges in adapting traditional editing techniques to modern DiT architectures. While our analysis centers on the memory footprint of attention score maps within a single timestep, editing methods based on attention engineering may involve caching or modifying attention maps across multiple timesteps, and we highlight the significant computation overhead when applying attention-engineering-based video editing methods to Video DiTs. As shown in Table~\ref{tab:attention_memory}, traditional diffusion models (e.g., Stable Diffusion and Zeroscope) exhibit multi-scale attention mechanisms with shapes varying by layer. In contrast, modern Video DiTs like CogVideoX-5B employ fixed large-scale attention ($[2,48,11490,11490]$), resulting in 283$\times$ higher memory than SD's maximum (7 GB vs. 1871 GB). This fundamental architectural shift explains the inefficiency of attention-based editing methods when applied to Video DiTs.
\begin{table}[ht]
    \label{tab:attention_memory}
    \centering
    \caption{\textbf{Peak attention memory consumption (GB) for full score maps (float32) per timestep.} Values with $\sim$ approximate ground truth with $\pm$5\% variance. $F$: processed frames. Asterisk (*) indicates dynamic attention shapes.}
    \label{tab:attention_memory}
    \resizebox{\linewidth}{!}{\begin{tabular}{lrrrrl}
    \toprule
        Model & Attention Shape & Block Number & Attention Memory (GB) & Dtype & $F$ \\
        \hline
        Stable Diffusion~\cite{rombach2022high}* & Multi-scale & 32 & \textasciitilde7 & float32 & 1 \\ 
        HunyuanDiT~\cite{li2024hunyuan} & [2,4096,4096] & 80 & \textasciitilde10 & float32 & 1 \\
        Zeroscope~\cite{wang2023modelscope}* & Multi-scale & 64 & \textasciitilde25 & float32 & 8 \\ 
        HunyuanVideo~\cite{kong2024hunyuanvideo} & [1,24,11520,11520] & 48 & \textasciitilde612 & float32 & 41 \\
        Wanx2.1-14B~\cite{wang2025wan} & [1,40,11264,11264] & 40 & \textasciitilde794 & float32 & 41 \\
        CogVideoX-5B~\cite{yang2024cogvideox} & [2,48,11490,11490] & 40 & \textasciitilde1871 & float32 & 41 \\ 
        \hline
    \end{tabular}}
\end{table}

\noindent\textbf{Inference Latency Comparison.} We adopt a theoretical estimation approach to evaluate the inference latency for attention-engineering-based editing methods (FateZero~\cite{qi2023fatezero} and KVEdit~\cite{zhu2025kv}) and measure the practical inference latency of DFVEdit, since direct empirical testing of FateZero and KVEdit with Video DiT is infeasible due to GPU memory and CPU RAM constraints. First, we conduct performance analysis assuming unlimited CPU memory. Specifically, we measure execution times per timestep and extrapolate to the total timesteps required for editing. Given that caching attention maps, keys, and queries exceeds GPU capacity, all caching operations utilize CPU memory. This methodology provides theoretical latency estimates for these methods in Video DiT contexts. The selected approaches demonstrate that both traditional diffusion-based methods and image DiT-based methods face significant resource overheads when directly applied to Video DiTs.

\subsection{More experiment results}
\noindent{\textbf{Embedding Reinforcement.} Due to space limitation, we only visualize the embedding reinforcement ablation results on stylization in the main text, here we additionally visualize the results on shape editing.
As shown in Fig.~\ref{fig:abltion_er_shape}, at $\gamma=0$, the distinctive traits of polar bears compared to brown bears, such as their white fur and rounded ears, are effectively captured with high background fidelity. Increasing $\gamma$ to 1 enhances editing quality, more accurately portraying the polar bear's elongated neck and smaller head-to-body ratio. However, at $\gamma=5$, there is a notable decline in video synthesis quality, characterized by visible flickering and noise, alongside reduced background preservation. Our experiments demonstrate that for optimal editing outcomes in shape modification, the ER method's hyperparameter $\gamma$ should be set within the range of 0 to 1. For simplicity and efficiency, we typically set $\gamma$ to 0.3 in our studies, although values within this range generally yield satisfactory results.

\noindent{\textbf{On multi-objects editing.}} Our method, though not specifically designed for multi-object editing, inherently adapts to such tasks through the editing region localization capability of CDFV. We enhance the editing precision by combining: (1) Implicit Cross-Attention (ICA) derived from Layer~16 transformer blocks (based on the observation that cross-attention masks exhibit a coarse-to-fine change across denoising timesteps as found in FreeMask~\cite{cai2024freemask}, with the layer index selection method also following FreeMask), and (2) SAM masks with edge padding. Fig.~\ref{fig:ICA} shows an example of ICA extraction and visualization. This strategy operates in two phases: ICA guidance during early denoising ($t=T\to0.4T$) preserves shape flexibility while reducing background leakage, followed by SAM-based mask guidance ($t=0.3T\to0$). 

As shown in Fig.~\ref{fig:multi_object}, our method demonstrates robust multi-object editing capabilities across diverse scenarios, achieving target object accuracy while maintaining non-edited region fidelity. The framework handles both complex dynamic interactions with multiple objects in cluttered environments, and fine-grained editing requiring precise motion retention. These findings not only underscore the robustness and effectiveness of our proposed method but also lay a solid foundation for future advancements in multi-object editing.

\noindent{\textbf{More results on attribute editing.}} Due to space limitations, we include the visualization results of attribute editing in the Appendix. As shown in Fig.~\ref{fig:attribute}, our method demonstrates satisfactory performance on attribute editing, enabling the natural and seamless integration of both added and removed small objects within existing scenes. 

\noindent{\textbf{More results on extension experiments.}} We have demonstrated additional results of applying DFVEdit to the Wan2.1 base model in the main text. To further objectively evaluate the generality and robustness of DFVEdit on Video DiTs, Figure~\ref{fig:compare_extension} compares the performance of the same video editing tasks across different base models using our method. This comparison aims to reveal the variations in outcomes due to differences in models and to verify the robustness and generality of our approach. The experimental results indicate that although different base models may lead to slight variations in the editing outcomes, overall, all edited videos align well with the target prompt, and the editing quality meets the expected standards. These findings reflect the high robustness and generalization ability of our proposed method. \textbf{\textit{Refer to the 'DFVEdit.mp4' in the supplementary material for the dynamic video display. The code will be public upon publication of this work.}}
\section{Limitations}
Fig.~\ref{fig:limitation} indicates limitations of our method. In zero-shot video editing, it is challenging to maintain full detail fidelity in non-edited regions. Although our method outperforms others in improving fidelity, achieving perfect detail fidelity remains difficult. Even with ICA guidance, some detail loss occurs. Additionally, for editing traits with large variations, such as transforming a bicycle into a car, our method cannot support significant shape variations. It is suitable only for shape editing tasks with small layout variations.

\begin{figure*}[htbp] 
  \centering 
  \includegraphics[width=1\linewidth]{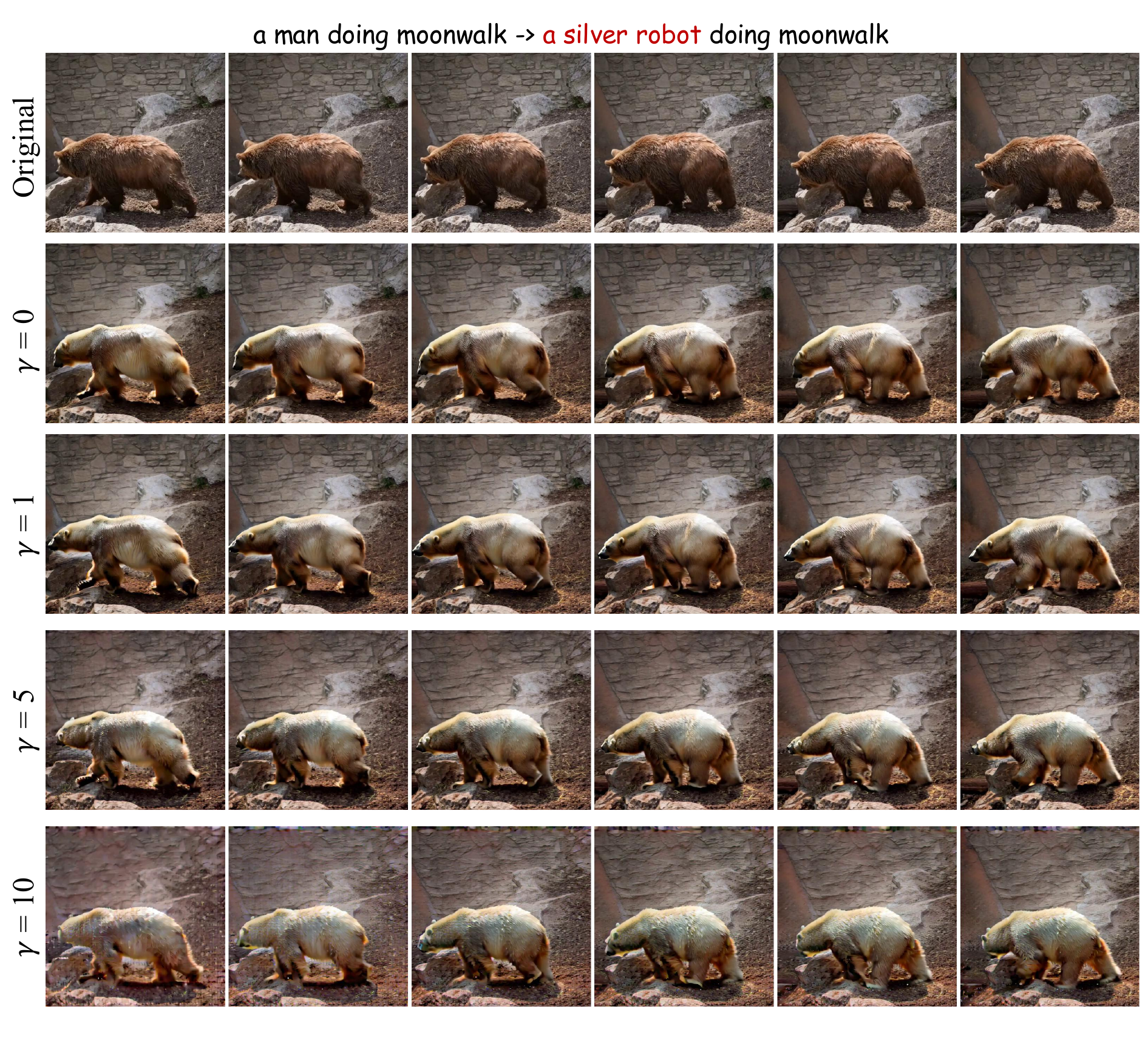} 
    \vspace{-1.5em}
    \caption{\textbf{Ablation results of ER on shape editing.} ER strength $\gamma$ ($\gamma=0\rightarrow1$ optimal, $\gamma\geq5$ degrades).}
  \label{fig:abltion_er_shape}
\end{figure*}

\begin{figure*}[htbp] 
  \centering 
  \includegraphics[width=1\linewidth]{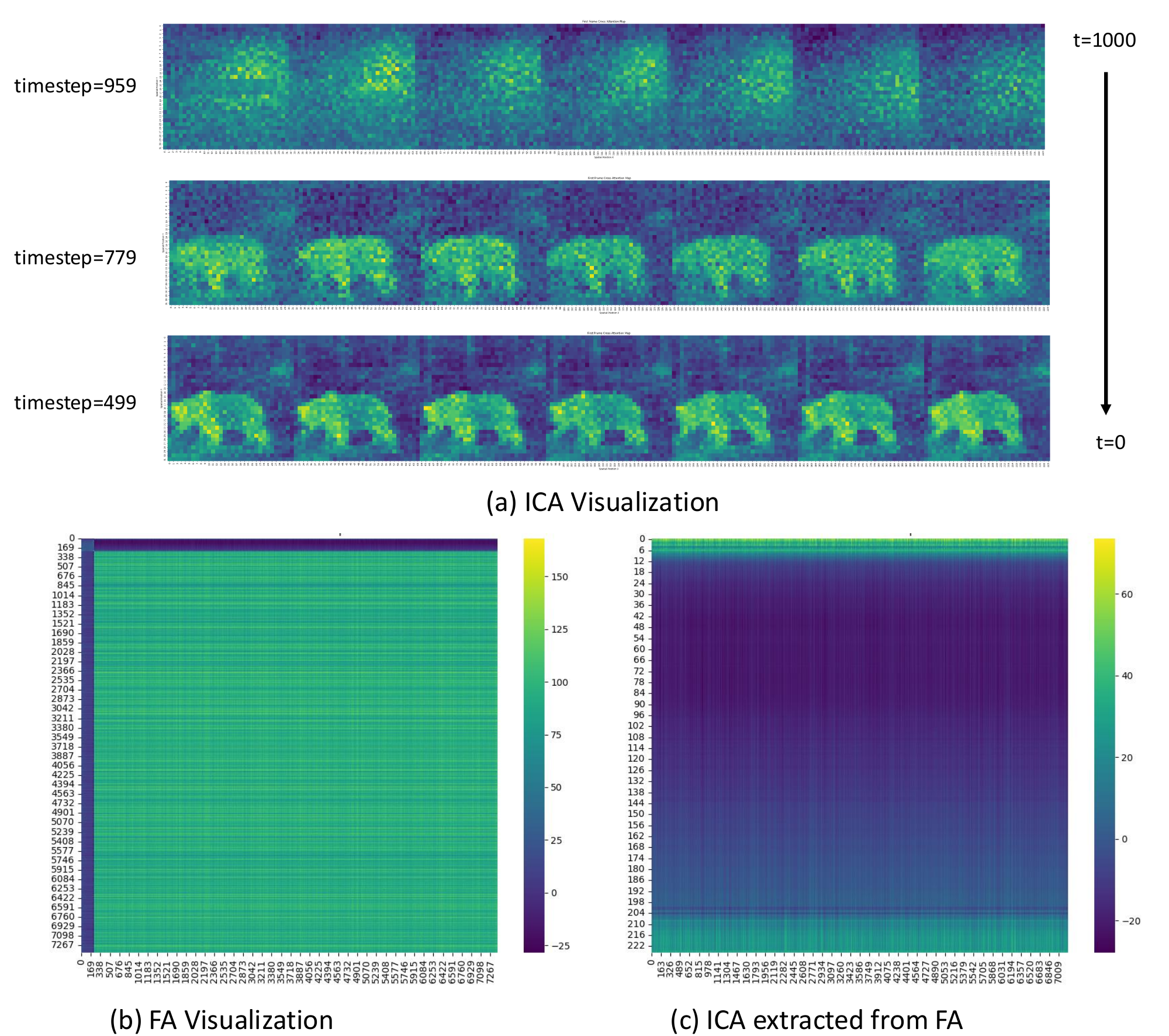} 
    \vspace{-1.5em}
    \caption{\textbf{Implicit Cross Attention extraction and visualization.}}
  \label{fig:ICA}
\end{figure*}

\begin{figure*}[htbp] 
  \centering 
  \includegraphics[width=1\linewidth]{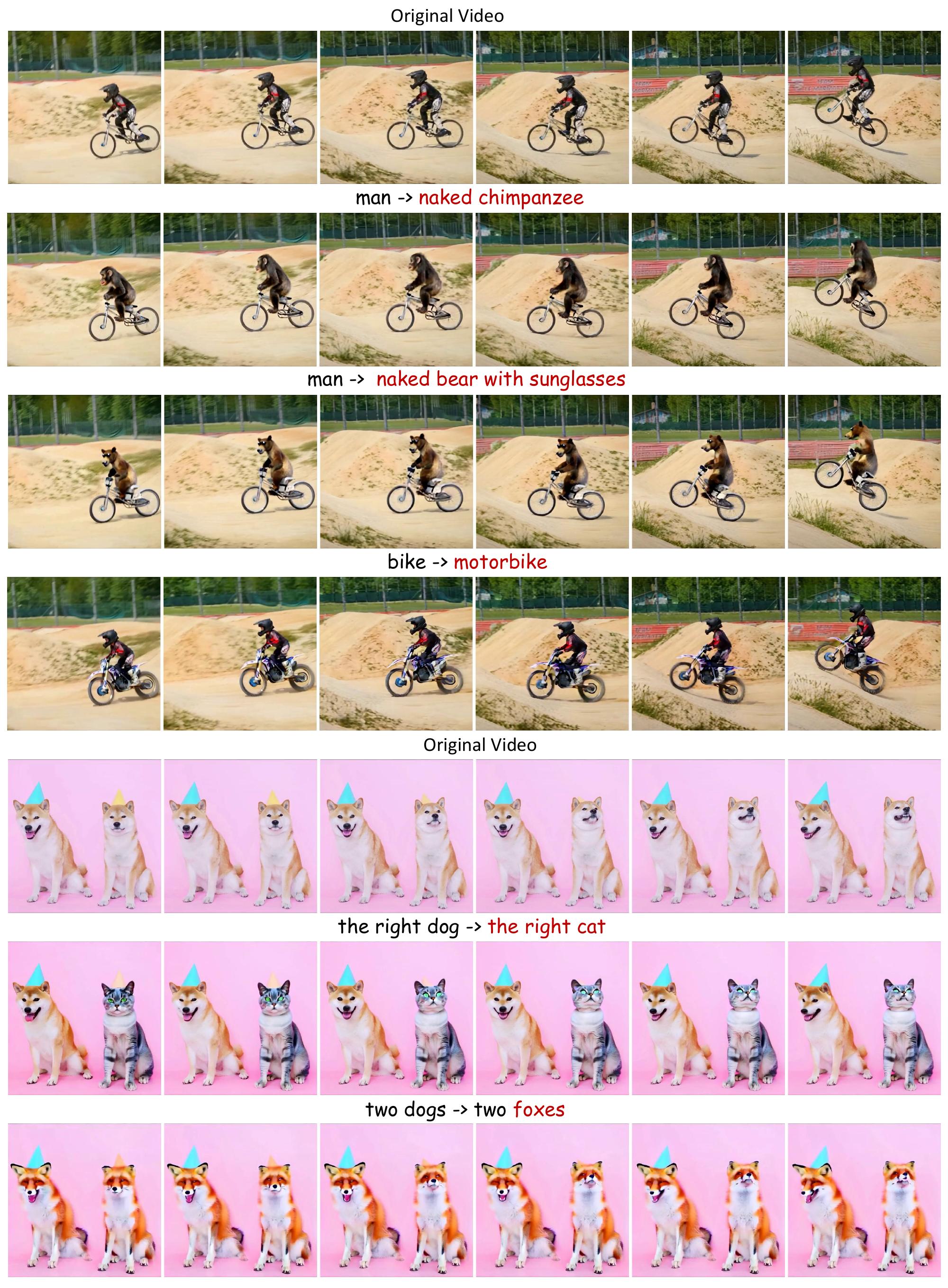} 
    \vspace{-1.5em}
    \caption{\textbf{Multi-object editing results.} DFVEdit performs well on multi-object editing across various scenarios: (1) complex dynamic interactions (person-vehicle) with cluttered backgrounds, and (2) fine-grained object manipulation with detailed motions (two dogs).}
  \label{fig:multi_object} 
\end{figure*}

\begin{figure*}[htbp] 
  \centering 
  \includegraphics[width=1\linewidth]{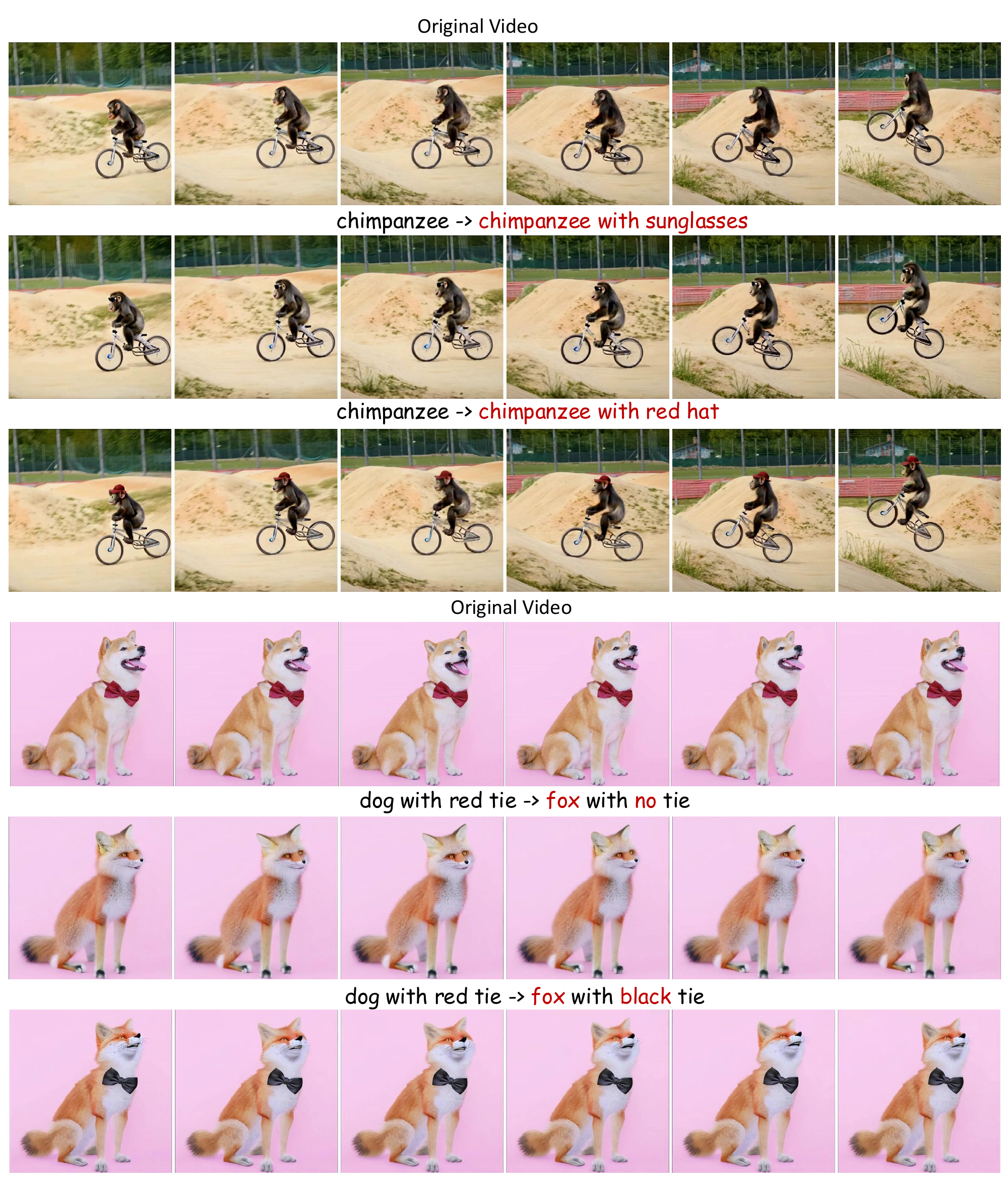} 
    \vspace{-1.5em}
    \caption{\textbf{Attribute editing results.}}
  \label{fig:attribute} 
\end{figure*}

\begin{figure*}[htbp] 
  \centering 
  \includegraphics[width=1\linewidth]{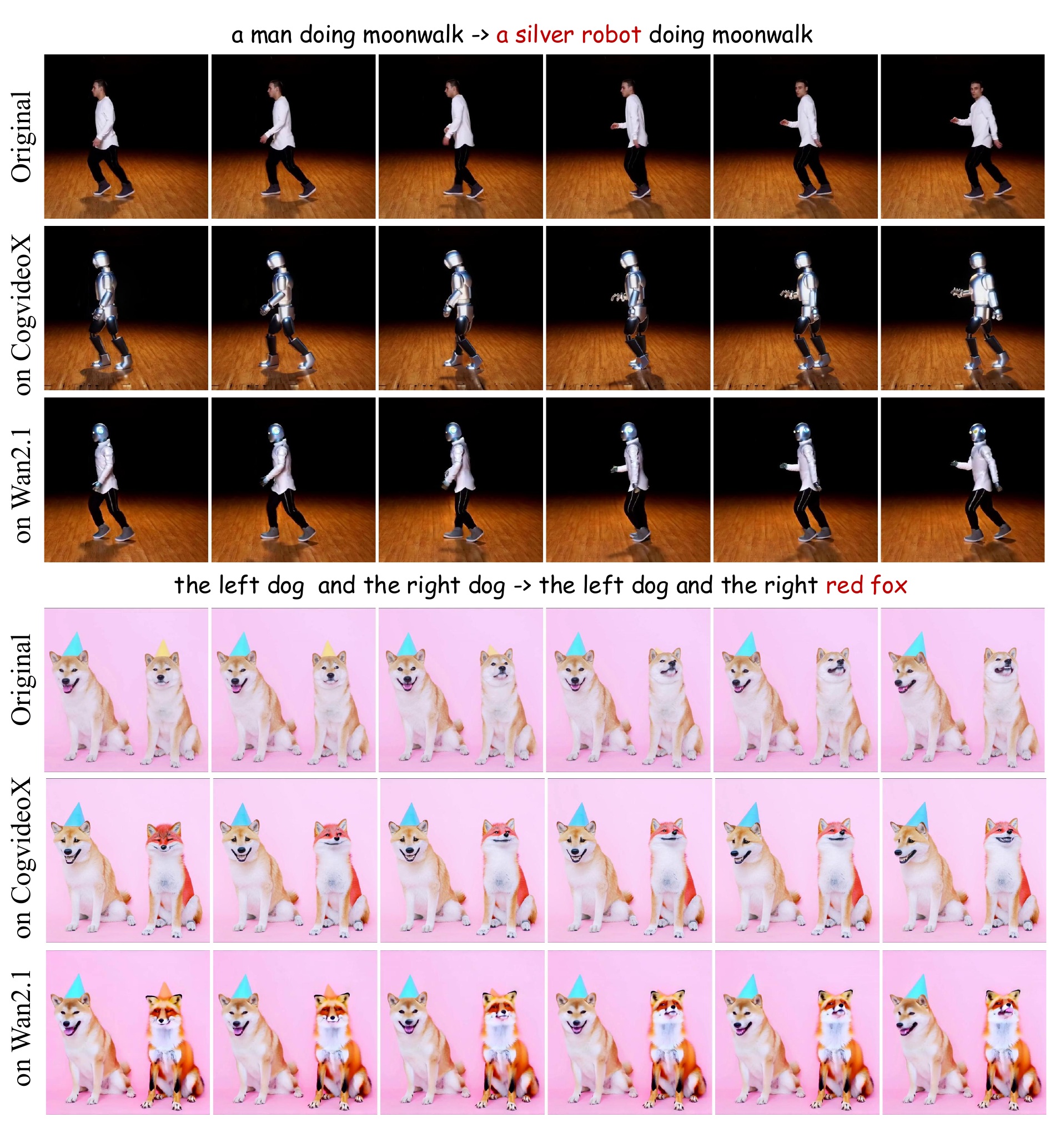} 
    \vspace{-1.5em}
    \caption{\textbf{More extension experiment results.}}
  \label{fig:compare_extension} 
\end{figure*}

\begin{figure*}[htbp] 
  \centering 
  \includegraphics[width=1\linewidth]{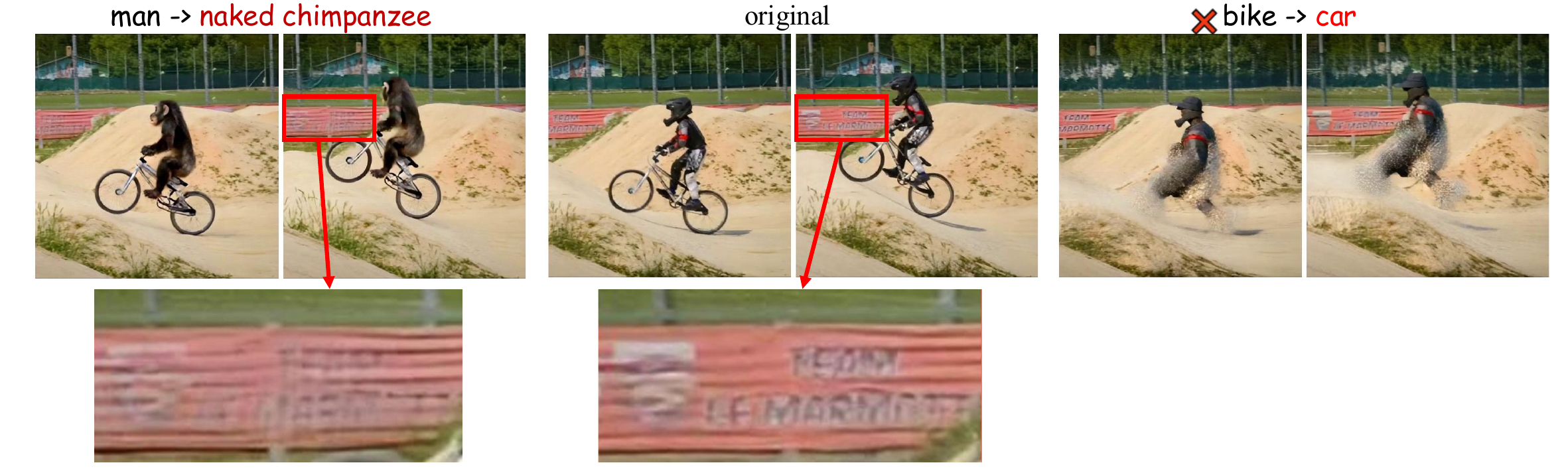} 
    \vspace{-1.5em}
    \caption{\textbf{Limitation.}}
  \label{fig:limitation} 
\end{figure*}

\clearpage

\end{document}